\documentclass{applemlr}

\usepackage{amsthm}

\theoremstyle{plain} % or 'definition' or 'remark' for different looks

\newtheorem{prop}{Proposition}[section]

\usepackage{todonotes}
\usepackage{algorithm}
\usepackage{algpseudocode} 
\usepackage{bbm}
\usepackage{dsfont}
\usepackage{amsmath,amsfonts,amssymb}

\usepackage{multirow}
\usepackage{graphicx}
\usepackage{subcaption} 

\usepackage{etoc}

\crefname{prop}{proposition}{propositions}
\Crefname{prop}{Proposition}{Propositions}

\usepackage{amsmath}
\usepackage{enumerate}
\usepackage{algorithm}
\usepackage{algpseudocode}
\usepackage{amsfonts}
\usepackage{amsthm}
\usepackage{cleveref}
\usepackage{diagbox}
\usepackage{colortbl}
\usepackage{amssymb}
\usepackage{xspace}
\usepackage{wrapfig}
\usepackage{adjustbox}
\usepackage{tabularx}
\usepackage{booktabs}
\usepackage{mathtools}
\usepackage{tikz}
\usepackage{enumitem}
\usepackage{silence}
\usepackage{dsfont}
\usepackage[table]{xcolor}
\usepackage[dvipsnames]{xcolor}
\usepackage{multirow}
\usepackage{makecell}
\usepackage{xfakebold}
\usepackage{amsmath,amsfonts,bm}

\def\eqref#1{equation~\ref{#1}}
\def\1{\bm{1}}

\DeclareMathAlphabet{\mathsfit}{\encodingdefault}{\sfdefault}{m}{sl}
\SetMathAlphabet{\mathsfit}{bold}{\encodingdefault}{\sfdefault}{bx}{n}

\definecolor{textgray}{HTML}{6E6E73}
\usetikzlibrary{positioning, calc}
\usetikzlibrary{decorations.pathmorphing}

\makeatletter
\patchcmd{\wrong@fontshape}{\@gobbletwo}{}{}{}
\makeatother
\numberwithin{equation}{section}
\makeatletter
\AtBeginDocument{
  \urlstyle{sf}
  
}
\makeatother

\definecolor{light}{RGB}{125, 125, 125}
\crefname{tcb@cnt@pbox}{code}{code}
\Crefname{tcb@cnt@pbox}{Code}{Code}
\crefname{assumption}{assumption}{assumption}
\Crefname{assumption}{Assumption}{Assumptions}

\newtcolorbox[auto counter]{pbox}[2][]{
  colback=white,
  title=Code~\thetcbcounter: #2,
  #1,fonttitle=\sffamily,
  fontupper=\sffamily,
  arc=2pt,
  colframe=bgcolor,
  coltitle=fgcolor,
  colbacktitle=bgcolor,
  toptitle=0.25cm,
  bottomtitle=0.125cm
}

\makeatletter
\newcommand\applefootnote[1]{%
  \begingroup
  \renewcommand\thefootnote{}%
  \renewcommand\@makefntext[1]{\noindent##1}%
  \footnote{#1}%
  \addtocounter{footnote}{-1}%
  \endgroup
}
\makeatother

\definecolor{cverbbg}{gray}{0.90}

\newcommand{\ourmethod}{FS-DFM\xspace}

\title{\ourmethod: Fast and Accurate Long Text \\ Generation with Few-Step Diffusion Language Model}

\author{%
  \textbf{Amin Karimi Monsefi\textsuperscript{1} \textsuperscript{‡},
    Nikhil Bhendawade \textsuperscript{2},
   Manuel R. Ciosici\textsuperscript{2},
   Dominic Culver \textsuperscript{2}},
   Yizhe Zhang\textsuperscript{2}, 
  Irina Belousova \textsuperscript{2}  \\
  \textsuperscript{1}The Ohio State University, 
  \textsuperscript{2}Apple \\
  \textsuperscript{‡}Work done during the internship at Apple
}

\abstract{
Autoregressive language models (ARMs) deliver strong likelihoods, but are inherently serial: they generate one token per forward pass, which limits throughput and inflates latency for long sequences. Diffusion Language Models (DLMs) parallelize across positions and thus appear promising for language generation, yet standard discrete diffusion typically needs hundreds to thousands of model evaluations to reach high quality, trading serial depth for iterative breadth. We introduce \textbf{\ourmethod}, Few-Step Discrete Flow-Matching. A discrete flow-matching model designed for speed without sacrificing quality. The core idea is simple: make the number of sampling steps an explicit parameter and train the model to be consistent across step budgets, so one big move lands where many small moves would. We pair this with a reliable update rule that moves probability in the right direction without overshooting, and with strong teacher guidance distilled from long-run trajectories. Together, these choices make few-step sampling stable, accurate, and easy to control. On language modeling benchmarks, \ourmethod with 8 sampling steps achieves perplexity parity with a 1\,024-step discrete-flow baseline for generating 1\,024 tokens using a similar-size model, delivering up to 128× faster sampling and corresponding latency/throughput gains.}

\metadata[Code]{\url{https://github.com/apple/ml-fs-dfm}}
\metadata[Correspondence]{\sffamily Amin Karimi Monsefi: \url{karimimonsefi.1@osu.edu}; Irina Belousova: \url{ibelousova@apple.com}}
\date{\sffamily\today}

\begin{document}

\maketitle

% \section{Introduction}

% \section{Regular Font Tests}
% Regular text: Testing -- (en-dash) and --- (em-dash) ligatures here.
% Also testing \textbf{bold text: Testing -- (en-dash) and --- (em-dash) ligatures here.}

% \section*{Direct Font Commands}
% {\fontfamily{sfpro}\selectfont Regular SF Pro: Testing -- and --- ligatures. Question mark: ?}

% {\fontfamily{sfpro}\bfseries\selectfont Bold SF Pro: Testing -- and --- ligatures. Question mark: ?}

% {\fontfamily{sfpro}\itshape\selectfont Italic SF Pro: Testing -- and --- ligatures. Question mark: ?}

\applefootnote{ \textcolor{textgray}{\sffamily Apple and the Apple logo are trademarks of Apple Inc., registered in the U.S. and other countries and regions.}}

\section{introduction}

\begin{figure}[h]
	\centering
	\includegraphics[width=1.\linewidth]{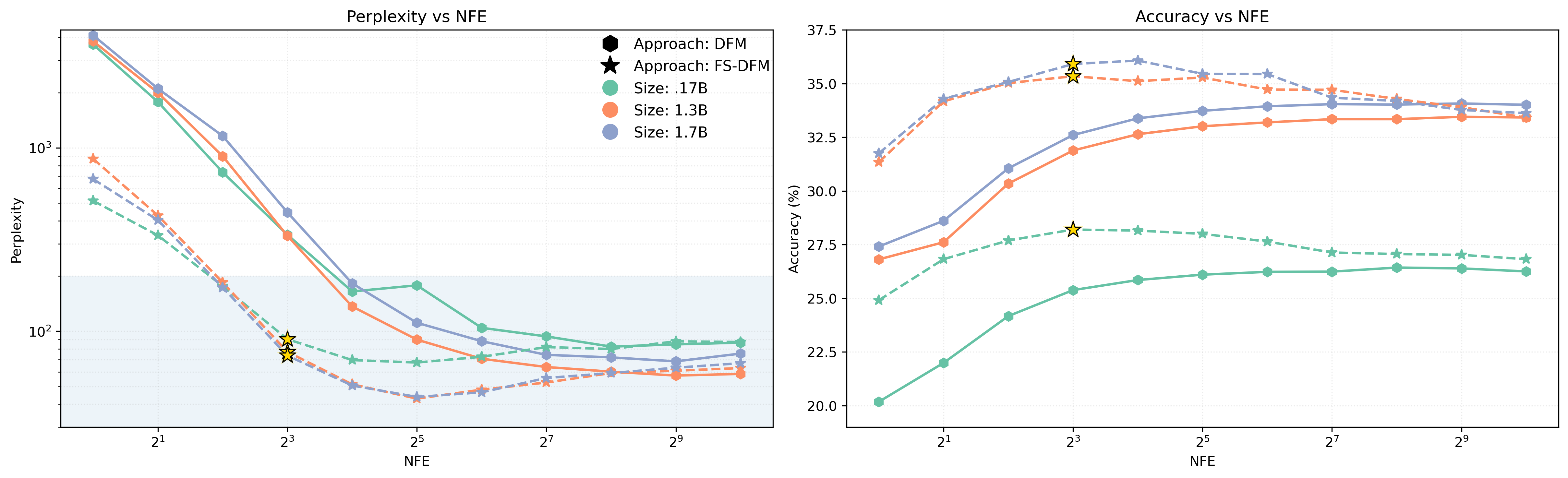}
	\vspace{-7mm}
	\caption{\small \textbf{Generation quality across model sizes} (perplexity and accuracy vs. NFE). \ourmethod reaches the strong-quality regime in few steps across all sizes, while DFM needs far more evaluations. Gold stars (NFE=8) highlight \ourmethod in a few-step regime, with accuracy quickly saturating and entropy converging to similar ranges as steps increase. The average value of entropy for all the models is $7.41$ to $8.07$.}
	\label{fig:fdm_vs_fs-fdm}\vspace{-3mm}
	    \label{fig:demo_fdm_fs-fdm}
\end{figure}

Autoregressive language models~(ARMs) generate sequences by predicting the next token conditioned on the observed prefix, and have achieved remarkable success~\citep{yang2025qwen3, team2023gemini, dubey2024llama, li2025apple}.  
In contrast, Diffusion Language Models~(DLMs) synthesize text through iterative refinement, providing stable likelihood-grounded training objectives, inherent parallelism across positions, and enhanced controllability through access to global context, capabilities that help mitigate exposure bias and naturally accommodate structure beyond left-to-right generation~\citep{nieLargeLanguageDiffusion2025, austin2021structured}. Despite these complementary advantages, both paradigms face fundamental bottlenecks. ARMs are constrained by sequential decoding, requiring one forward pass per token, which limits throughput and enforces a unidirectional dependency that hinders tasks such as reversal or order-invariant reasoning~\citep{zhao2023survey, schulman2022chatgpt, gong2025diffucoder}. Although DLMs parallelize over positions, they typically require tens to hundreds of refinement steps to achieve competitive quality, effectively replacing the autoregressive depth of a single long pass with a stack of iterative model evaluations~\citep{chen2023learning, arriolaBlockDiffusionInterpolating2025, nie2024scaling}. For example, to march the ARM generation quality, LLaDA requires roughly one inference step per output token~\citep{nieLargeLanguageDiffusion2025}.

\begin{figure}[t]
    \centering
    \includegraphics[width=0.95\linewidth]{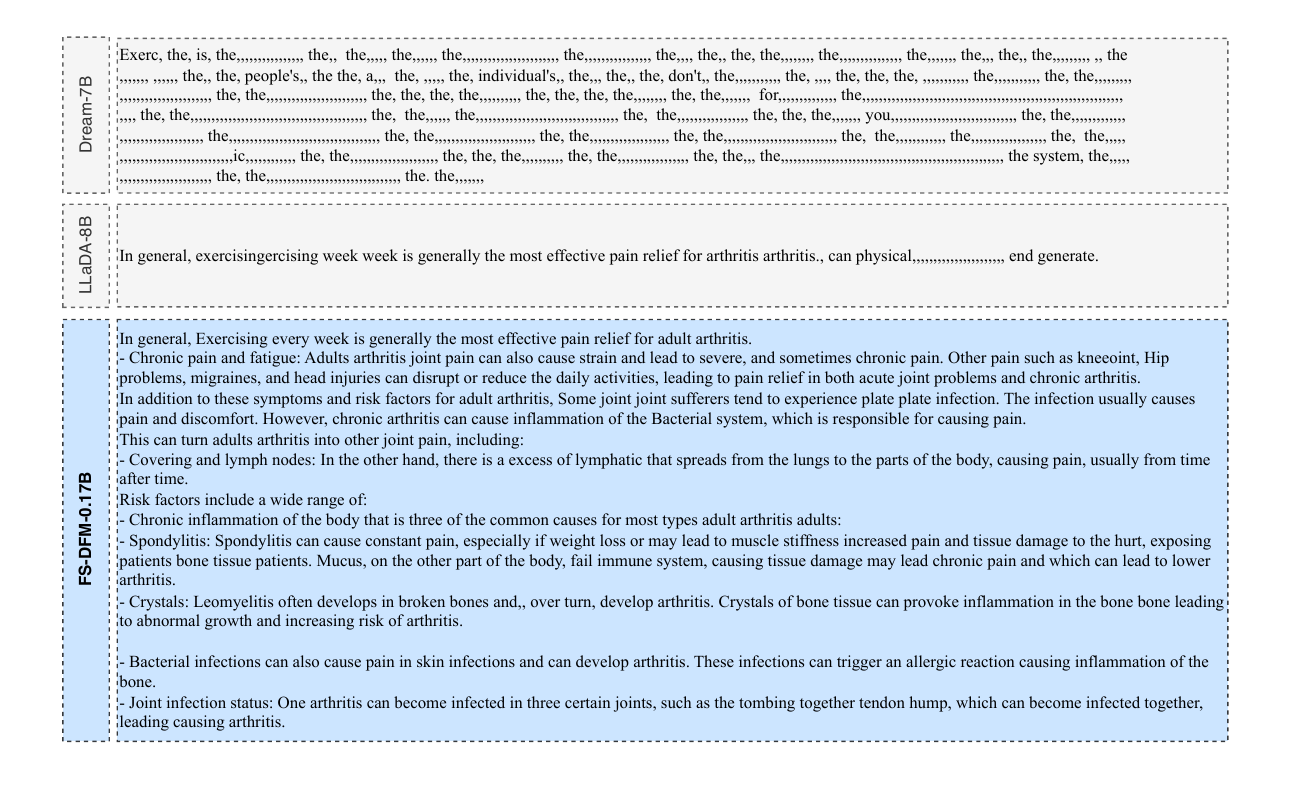}
    \vspace{-7mm}
    \caption{\small
    \textbf{Eight-step long-horizon generation:}
    1\,024-token unconditional generation in \emph{8 sampling steps}.
    \ourmethod (0.17B) successfully produces 1\,024 tokens under the 8-step constraint. Despite having 40x more parameters, LLaDA-8B-Instruct and Dream-7B-Instruct's 8-steps generations exhibit trailing blanks and punctuation artifacts (e.g., repeated commas).
    Generations are truncated. Complete output in \Cref{app:sample_outputs}.
    }
    \label{fig:demo}\vspace{-3mm}
\end{figure}

Continuous-space flow-matching models have emerged as a streamlined alternative to conventional score-based diffusion trained around reverse Stochastic Differential Equations~(SDEs). Instead of modeling a stochastic reverse process, flow-matching directly trains a continuous normalizing flow via vector-field regression, enabling deterministic probability-flow Ordinary Differential Equation~(ODE) sampling. This approach often achieves comparable or superior likelihoods and sample quality with fewer function evaluations and more stable optimization (e.g., through optimal transport or straight-line paths) than SDE-based pipelines~\citep{lipman2022flow, song2020score, liSurveyDiffusionLanguage2025, gong2022diffuseq}.  
Discrete Flow-Matching~(DFM) extends these advantages to discrete domains by adapting flow-matching to tokenized data, providing flexible probability paths and strong perplexity at scale. However, DFM still requires a substantial number of evaluations along each trajectory, hindering applicability~\citep{NEURIPS2024_f0d629a7}.

We introduce Few-Step Discrete Flow-Matching~(\ourmethod), a new diffusion-based language model built upon the DFM paradigm, which preserves the key advantages of prior diffusion and flow models (bidirectional context utilization and flexible generation) while explicitly mitigating speed and sampling overhead weaknesses. \ourmethod uses DFM's discrete-flow perspective to substantially reduce the number of refinement steps required for high-quality generation. In doing so, \ourmethod narrows the inference efficiency and modeling performance gap between DLMs and ARMs. \ourmethod can reach the quality regime of a 1\,024-step DFM with just 8 NFEs (Number of Function Evaluations), 128× fewer steps for comparable quality (\Cref{fig:demo_fdm_fs-fdm}).

\ourmethod builds on two pillars. First, we enrich the model with the shortcut principle \citep{frans2024one} together with a self-consistency constraint: the network is conditioned not only on masked tokens but also on the desired step size, and trained so that a single large step agrees with the composition of multiple smaller steps. We instantiate the shortcut teacher using a Runge–Kutta formulation —we evaluate RK-2 (Heun) and classical RK-4 as ODE estimators— and adopt the one that offers the best accuracy–stability trade-off for large single/few-step updates. Second, a DFM reformulation that models sequences as a Continuous-Time Discrete Markov Chain (CTMC, \citet{campbell2022continuous}) and a probability velocity is learned to transport a source distribution to the data distribution along a probability path~\citep{NEURIPS2024_f0d629a7, campbell2024generative}. In this framework, a velocity field advances samples ``locally'' from $p_t$ to $p_{t+h}$. Instead, \ourmethod learns a \emph{Cumulative Scalar} parameterized by the step size $h$ to enable large and reliable jumps along the probability path. The step-budgeted perspective retains the core benefits of flow-matching: stable training, parallel refinement, and bidirectional context, while elevating few-step generation to a first-class design objective.

To our knowledge, \ourmethod is the first few-step discrete flow–matching approach aimed at long-horizon language modeling. Prior discrete diffusion/flow methods require hundreds or even thousands of refinement steps, often growing with sequence length~\citep{NEURIPS2024_f0d629a7, nieLargeLanguageDiffusion2025}. \ourmethod works directly in token space with step-aware DFM and a closed-form cumulative scalar update, matching the perplexity of a 1\,024-step discrete-flow baseline for 1\,024-token generation in just 8 steps—up to \textbf{128$\times$} \textbf{faster while preserving long-horizon quality.}

\section{Related Work}

Diffusion models have demonstrated remarkable effectiveness in continuous domains such as images, achieving state-of-the-art quality across diverse tasks~\citep{karimi2025taxadiffusion, zhang2023adding, navard2024knobgen, rombach2022high}. Extending diffusion-style generative modeling to text, Diffusion Language Models (DLMs), have shown impressive task performance coupled with fast inference through parallel generation~\citep{nieLargeLanguageDiffusion2025,yeDream7BDiffusion2025,labsMercuryUltraFastLanguage2025}. DLMs can be broadly grouped into  
\emph{continuous} models, which denoise in embedding or latent space, and \emph{discrete} models, which operate directly over tokens. Both categories exhibit the key advantages of diffusion, including bidirectional context, controllability, and parallel refinement~\citep{liSurveyDiffusionLanguage2025}. 

Continuous DLMs~\citep{li2022diffusion, lin2023text, lovelace2023latent} cast text generation as iterative denoising in a continuous embedding or latent space. A forward corruption process perturbs token embeddings or hidden states, while a parameterized denoiser reconstructs clean representations conditioned on the context. Continuous DLMs take advantage of benefits that arise naturally in continuous representations: flexible conditioning and guidance, bidirectional contextualization during refinement, and compatibility with large-scale pretraining or task-specific fine-tuning. Moreover, continuous DLMs benefits from research in diffusion models for other modalities such as images or audio as methods easily port between domains.

Discrete DLMs, on the other hand, model token dynamics either through discrete-time noise schedules or continuous-time Markov formulations. DLMs such as LLaDA~\citep{nieLargeLanguageDiffusion2025} and Dream~\citep{yeDream7BDiffusion2025} match autoregressive models in task performance if allowed sufficient refinement steps. Effectively, they tie step count to sequence length and generation quality~\citep{austin2021structured, gong2024scaling, zheng2023reparameterized, shi2024simplified, ou2024your}. One line of research accepts this high step count requirement and focuses on speeding up inference through approaches such as confidence-aware parallel decoding, tweaks to re-enable the KV Cache, and restricted attention~\citep{wuFastdLLMTrainingfreeAcceleration2025,arriolaBlockDiffusionInterpolating2025,chenDPadEfficientDiffusion2025}. Another line of research focuses on reducing the number of inference steps through one-/few-step approaches~\citep{NEURIPS2024_f0d629a7}. Our work falls within this second line of research.

Prior work on one-step language generation primarily focused on continuous-space diffusion models, operating over embeddings rather than discrete tokens. DLM-One~\citep{chen2025dlm} demonstrated large speedups on short sequence tasks but did not address long sequence modeling. Similarly, FlowSeq~\citep{hu2024flow} operates on continuous embeddings, bypassing token-level likelihoods. Within the discrete family, recent methods typically attain quality with thousands of refinement steps, often scaling with sequence length~\citep{NEURIPS2024_f0d629a7,zhao2023survey}. SDTT~\citep{deschenaux2024beyond} distills discrete DLMs for faster sampling, yet still uses \mbox{16–256} steps and focuses on short-form tasks. In contrast, our method (\ourmethod) operates \emph{in the discrete space} and explicitly targets \emph{the few-step regime} (1-8 steps). \ourmethod can match 1\,024-step discrete-flow baselines in just 8 steps while preserving text quality.

\section{Preliminaries and Background}
\label{sec:prelim}

We provide a brief summary on flow-matching, following \citet{NEURIPS2024_f0d629a7}. For the notation used and for details on the theory behind discrete flow-matching, readers are referred to \Cref{appendix:preliminaries}.

\paragraph{Setup.}

Fix a target distribution $p_1$ of text, thought of as length $L$ sequences of token ids. The starting point of the flow-matching approach to text generation is a probability path $(p_t)_{0 \leq t \leq 1}$ that interpolates between a source distribution $p_0$ and the target distribution $p_1$. Given this path, we can sample $X_t \sim p_t$ for $t\in [0, 1]$, which leads to a stochastic process $(X_t)_{0\leq t \leq 1}$. 

Discrete Flow-Matching (DFM) models the stochastic process $(X_t)_{0\leq t \leq 1}$ as a Continuous-Time Markov Chain (CTMC). The evolution of this CTMC can be captured by its \emph{infinitesimal generator} $u_t(\cdot, \cdot)$ for $t\in [0, 1]$. For token sequences $x, y$, the value $u_t(x, y)$ captures the rate of change of the probability that the sequence will change from state $x$ to state $y$ and must satisfy the conditions in \Cref{prop:app-q-matrix-conditions}. We also have the following equality for small $h$ (see \Cref{appendix:notation} for notation)
\begin{equation}
    \mathbb{P}(X_{t+h} = y | X_t = x) = \delta_{x}(y) + h u_t(x, y) + o(h).
\end{equation}
The goal of DFM is to learn the infinitesimal generator $u_t$. DFM makes the further simplifying assumption (to reduce output dimension size) that $u_t(x, y)$ can be factorized as
\begin{equation}\label{eqn:factorized-velocities}
    u_t(x, y) = \sum_i \delta_{\overline{x^i}}(\overline{y^i})u_t^i(y^i, x).
\end{equation}
Similarly, the token-level $u_t^i$ must satisfy analogous conditions to \Cref{prop:app-q-matrix-conditions}. This reduces the output dimension from $|V|^L$ to $|V|\cdot L$, making the learning objective feasible. From this, we then have, for small $h$,
\begin{equation}
    \mathbb{P}(X_{t+h}^i = y^i | X_t = x) = \delta_{x^i}(y^i) + hu_t^i(y^i, x) + o(h).
\end{equation}
This reduces DFM to learning the factors $u_t^i$. It also provides a way of simulating the CTMC and generating samples from noise; given a sample $X_t$ we sample $X_{t+h}^i$ via Euler sampling
\[
X_{t+h}^i \sim \delta_{X_t^i} + h\cdot u_t^i(\cdot, X_t) + o(h), \quad i = 1, \ldots, L
\]

Of course, this depends on the choice of probability path $(p_t)_{0\leq t \leq 1}$. Following \cite{NEURIPS2024_f0d629a7}, start with
\[
p_t(x) = \sum_{x_0, x_1}p_t(x | x_0, x_1)\pi(x_0, x_1)
\]
where $x_0 \sim p_0$, $x_1 \sim p_1$, and $\pi$ is a joint distribution which relates $p_0$ and $p_1$\footnote{Namely, $\sum_{x_0} \pi(x_0, x_1) = p_1(x_1)$ and $\sum_{x_1} \pi(x_0, x_1) = p_0(x_0)$}, and
\[
p_t(x | x_0, x_1) = \prod_i p_t(x^i | x_0, x_1).
\]
Then set the conditional probabilities $p_t(x^i | x_0, x_1)$ to be the convex sum:
\begin{equation}\label{eqn:conditional-probability-path}
    p_t(x^i | x_0, x_1) := (1-\kappa_t)\delta_{x_0}(x^i) + \kappa_t\delta_{x_1}(x^i),
\end{equation}
so that the probability path of the $i$th token position is a linear interpolation between the source and the target distributions. Here, $\kappa_t$ is known as a \emph{scheduler} and must be a monotonically increasing differentiable function $\kappa: [0, 1] \to [0, 1]$ where $\kappa_0 = 0$ and $\kappa_1 = 1$. Using the ``Marginalization Trick'', one can derive the following description of the factorized velocities $u_t^i$
\begin{equation}\label{eqn:marginal-velocity}
    u_t^i(x^i, z) = \frac{\dot{\kappa_t}}{1-\kappa_t}\left(p_{1|t}(x^i|z) - \delta_z(x^i)\right)
\end{equation}
where $z$ is a token sequence and
\begin{equation}
    p_{1|t}(x^i|z) := \sum_{x_0, x_1}\delta_{x_1}(x^i)p_t(x_0, x_1 | z).
\end{equation}
In DFM, the model only needs to learn $p_{1|t}(x^i|z)$. For notational convenience, we also set
\begin{equation}\label{eqn:defn-g}
    g(t) := \frac{\dot{\kappa_t}}{1-\kappa_t}.
\end{equation}

\textbf{Learning the Denoiser.}
Given the factorized velocities derived above, the remaining task is to learn $p_{1|t}(\cdot|z)$. We parameterize this conditional distribution using a network $\theta$ that outputs logits:
\[
p_{1|t}(x^i | z) = \mathrm{softmax}(\theta^i(z, t))
\]
where $\theta^i(z, t)$ denotes the logits for the $i$-th token position.

To train this model, we need an appropriate loss function. We use the Bregman divergence for the loss function \cite[Equation 7.31]{lipman2024flow}. Starting from the velocity formulation in \Cref{eqn:marginal-velocity}, we can construct a loss that encourages the model to correctly predict $p_{1|t}$. Given a sample trajectory where $x_0 \sim p_0$, $x_1 \sim p_1$ are related through $\pi$, and $x_t$ is sampled from $p_t(x | x_0, x_1)$, the per-token loss at position $i$ is:
\begin{equation}\label{eq:dfm:loss}
   \mathcal{L}_i(x_1, x_t, t) = -g(t)\left[
p_{1|t}(x_t^i | x_t) - \delta_{x_1^i}(x_t^i) + \left(1-\delta_{x_1^i}(x_t^i)\right)\log p_{1|t}(x_1^i | x_t)\right]
\end{equation}
This loss encourages the model to assign high probability to the true target token $x_1^i$ when denoising from $x_t$. The scaling factor $g(t)$ naturally arises from the velocity formulation and ensures proper weighting across different time steps.

% To train this model, we need an appropriate loss function. Starting from the velocity formulation in \cref{eqn:marginal-velocity}, we can construct a loss that encourages the model to correctly predict $p_{1|t}$. Given a sample trajectory where $x_0 \sim p_0$, $x_1 \sim p_1$ are related through $\pi$, and $x_t$ is sampled from $p_t(x | x_0, x_1)$, the per-token loss at position $i$ is:
% \begin{equation}\label{eq:dfm:loss}
%    \mathcal{L}_i(x_1, x_t, t) = -g(t)\left[
% p_{1|t}(x_t^i | x_t) - \delta_{x_1^i}(x_t^i) + \left(1-\delta_{x_1^i}(x_t^i)\right)\log p_{1|t}(x_1^i | x_t)\right]
% \end{equation}
% This loss encourages the model to assign high probability to the true target token $x_1^i$ when denoising from $x_t$. The scaling factor $g(t)$ naturally arises from the velocity formulation and ensures proper weighting across different time steps.

\section{Method}
\label{sec:method}

Our approach comprises two components: \emph{Step-Aware Discrete Flow-Matching} and a \emph{Cumulative Scalar} update.
The first component exposes the step budget \(h\) as an explicit control signal within the DFM generator $u_t$ and distills from a shortcut teacher model~\citep{frans2024one}, allowing the learner to make large, single/few-step moves that approximate the cumulative effect of many small updates. The second is using a cumulative scalar to aid the model in jumping from a source token to a target token with large single/few-step predictions.

\subsection{Step-Aware Discrete Flow-Matching}
\label{sec:method:subsec:step-Aware}

We expose the \emph{step budget} \(h\) as an explicit conditioning signal to the DFM generator so the model learns transitions calibrated to the intended number of sampling steps. To expose the step budget to the model during training, we need a way to produce the Markov chain transition probabilities over large step intervals. However, determining the transition probabilities from the generator $u_t$ is intractable. But, because the transition probabilities satisfy the Kolmogorov equations (\Cref{eqn:kolmogorov-forward,eqn:kolmogorov-backward}) which are ODEs, we can use numerical methods to approximate the transition probabilities. The role of the teacher model is to approximate the evolution of the Markov chain, to be used as the ground truth for training \ourmethod. Many methods exist for approximating the ODE, each with different trade-offs between accuracy and computational efficiency. We experimented with two Runge-Kutta methods: RK-2 (Heun average) and RK-4 and found that RK-4 greatly improved the results (\Cref{fig:rk_4_vs_heun}), though at the cost of being more computationally expensive. 

\textbf{RK-4 ODE solver.}
The RK-4 algorithm is presented in \Cref{alg:rk4-shortcut}. Define $\mathrm{Vel}$ to be the function which computes the velocity from the logits. In particular, given logits $\ell$, current state $x_t$, $\mathrm{Vel}$ computes the result of \Cref{eqn:marginal-velocity}:
\[
\mathrm{Vel}(\mathrm{softmax}(\ell), x_t, t) := g(t)\left(p_{1|t}(\cdot|x_t) - \delta_{x_t}(\cdot)\right).
\]
Given the velocity $u$ from $\mathrm{Vel}$, we apply jump sampling (see \Cref{app:sec:jump:chains}) to obtain $x = \mathrm{Jump}(x_t, u, h)$ where $h$ is our desired time step. For a given time $t$ and time step $h$, RK-4 computes the model logits at $t$ and $t+h/2$ using $\mathrm{Vel}$ and $\mathrm{Jump}$ as defined above. Note that we use $\mathrm{Jump}$ only to obtain the relevant state at $t + h/2$ in order to compute the logits there. RK-4 then averages these logits. We include an detailed explanation of RK-2 in \Cref{app:sec:rk-2-heun}.

\begin{algorithm}
\caption{Shortcut RK-4}
\label{alg:rk4-shortcut}
\begin{algorithmic}[1]
\Require tokens $x_t$, time $t$, step $h$; Model $\theta$; velocity $\mathrm{Vel}$; CTMC jumper $\mathrm{Jump}$; \textit{use\_ema} flag
\State $h' \gets h/2$;\quad $t_{\text{mid}} \gets t + h'$;\quad $t_{\text{next}} \gets t + h$
\State $\theta' \gets \text{EMA}(\theta)$ if \textit{use\_ema} else $\theta$
\State $\ell_1 \gets \theta'(x_t, t; h')$;\quad \ \ \ \ \ \ \ $u_1 \gets \mathrm{Vel}(\mathrm{softmax}(\ell_1), x_t, h', t)$;\quad \ \ \ \ \ \ \ \ $x^{(1)} \gets \mathrm{Jump}(x_t, u_1, h')$
\State $\ell_2 \gets \theta'(x^{(1)}, t_{\text{mid}}; h')$;\quad $u_2 \gets \mathrm{Vel}(\mathrm{softmax}(\ell_2), x^{(1)}, h', t_{\text{mid}})$;\quad $x^{(2)} \gets \mathrm{Jump}(x^{(1)}, u_2, h')$
\State $\ell_3 \gets \theta'(x^{(2)}, t_{\text{mid}}; h')$;\quad $u_3 \gets \mathrm{Vel}(\mathrm{softmax}(\ell_3), x^{(2)}, h', t_{\text{mid}})$;\quad $x^{(3)} \gets \mathrm{Jump}(x^{(2)}, u_3, h')$
\State $\ell_4 \gets \theta'(x^{(3)}, t_{\text{next}}; h')$
\State \textbf{RK-4 average:}\; $\bar{\ell} \gets \tfrac{1}{6}(\ell_1 + 2\ell_2 + 2\ell_3 + \ell_4)$
\State \Return $\bar{\ell}$
\end{algorithmic}
\end{algorithm}

\textbf{EMA teacher for stability.}
Because training is non-stationary, using the \emph{current} student as the teacher causes the shortcut target to drift with parameter updates, which destabilizes large steps \(h\) where local errors accumulate across sub-evaluations.
We therefore maintain a slowly varying exponential moving average~(EMA) teacher:
\begin{equation}
\theta' \leftarrow \beta\,\theta' + (1-\beta)\,\theta,\qquad \beta\in[0,1)\
\label{eq:ema}
\end{equation}
Stop gradients through \(\theta'\).
The EMA teacher provides stable, low-variance targets over \([t,t{+}h]\), improving convergence for large \(h\) and making self-consistency training robust across step budgets.

\subsection{Cumulative Scalar}
\label{sec:method:subsec:average-velocity}
For a CTMC token update, the marginal velocity separates into a \emph{scale} and a \emph{direction} (\Cref{eqn:marginal-velocity}):
\begin{equation}\label{eqn:marginal-velocity_2}
u_t^i(x^i,z)
\;=\;
\underbrace{\frac{\dot{\kappa}(t)}{1-\kappa(t)}}_{\text{scale } g(t)}
\cdot
\underbrace{\big[p_{1|t}(x^i\!\mid\!z)-\delta_z(x^i)\big]}_{\text{direction}}.
\end{equation}

\textbf{Scale and jump behavior:} With a monotone scheduler $\kappa:[0,1]\!\to\![0,1]$ ($\kappa(0)=0,\ \kappa(1)=1$), few/one-step sampling uses a large step $h$, so the first update typically occurs at small $t$; for common schedulers, this makes the instantaneous \emph{scale} $g(t)=\dot\kappa(t)/(1-\kappa(t))$ too weak to trigger moves, stalling in early steps (\Cref{app:checkerboard-demo} contains more details about this argument and our motivations). Empirically, the mean jumps per token are \(\approx1.05\) with a \emph{uniform} source and \(\approx1.00\) with a \emph{mask} source (most positions change at most once), so the direction term in \Cref{eqn:marginal-velocity_2} is typically ``spent'' in a single decisive update, after which tokens rarely move. In this regime, getting the scale right for each finite step dominates quality.

\begin{center}
\emph{\textbf{Key question:} How can we incorporate the current time t and the step budget h into the scale so that, even when t is small, a single finite step delivers the right amount of flow?}
\end{center}

To provide the correct amount of probability flow over a finite step, we replace the instantaneous scale by a \emph{Cumulative Scalar} obtained by integrating \(g\) over the interval and normalizing by its length:
\begin{equation}
G_{t,h}=\!\int_t^{t+h}\!\frac{\dot\kappa(\tau)}{1-\kappa(\tau)}\,d\tau
= \ln\!\frac{1-\kappa(t)}{1-\kappa(t{+}h)}\,,
\qquad
\bar g_{t,h}=\frac{G_{t,h}}{h}
=\frac{1}{h}\ln\!\frac{1-\kappa(t)}{1-\kappa(t{+}h)}.
\label{eq:sample:average:velocity}
\end{equation}
Substituting the Cumulative Scalar yields
\begin{equation}\label{eqn:marginal-velocity-with-scaler}
\bar u_t^i(x^i,z)=\bar g_{t,h}\,\big(p_{1|t}(x^i\!\mid\!z)-\delta_z(x^i)\big),
\end{equation}
which calibrates the step strength using both \(t\) and \(h\), enabling effective jumps even when \(t\) is small. Using the Cumulative Scalar in \Cref{eqn:marginal-velocity-with-scaler} addresses this and improves few-/one-step generation.

\subsection{Training Approach}
\label{sec:method:training}
We train the step-aware generator \(\theta\) to be \emph{locally faithful} to the DFM path at small steps and \emph{globally consistent} with a shortcut teacher over large steps. 
Each minibatch provides \((x_t, x_1, t, h)\), where \(h\in(0,1]\) and \(h + t \leq 1\) is the intended step size (the step budget). 
The student produces logits \(\ell=\theta(x_t, t; h)\), which are used both in a small-step \emph{path} objective (\Cref{eq:dfm:loss}) and for comparison against a large-step \emph{teacher} defined on \([t, t{+}h]\). The sampling process is detailed in \Cref{sec:method:sampling}.

\textbf{Shortcut teacher.}
To stabilize and supervise large moves, we integrate \(\theta\) across the interval using a shortcut scheme, implemented with an EMA copy \(\theta'\) yielding averaged logits \(\ell_{\text{tea}}\):
\[
\ell_{\text{tea}}
\;\leftarrow\; \text{RK-2 / RK-4 \ Estimate} \big(\theta', x_t, t, h\big)
\]
The teacher is treated as a stop–gradient (no backprop through \(\theta'\)).
After each optimizer step, we update the EMA parameters via ~\Cref{eq:ema}.

\textbf{Losses.}
Let $p_\theta(\cdot\mid x_t,t,h)=\mathrm{softmax}(\ell/T)$ and $p_{\text{tea}}(\cdot\mid x_t,t,h)=\mathrm{softmax}(\ell_{\text{tea}}/T)$ with temperature $T\ge1$ (in our experiments, $T$ is always set to $1$). Let $L$ denote the context length. 
We compute losses tokenwise over all positions and average per sample.
\[
 \mathcal{L}_{\text{dist}}
\;=\;
\frac{1}{L}
\sum_{j=1}^L  \,
D_{\mathrm{KL}}\!\big(p_{\text{tea},j}\,\|\,p_{\theta,j}\big)
\quad\text{(stop-grad on } \ell_{\text{tea}}\text{).}
\]
For the DFM path objective, we use the per-token loss from \Cref{eq:dfm:loss} averaged across tokens:
\[
\mathcal{L}_{\text{dfm}}
\;=\;
\frac{1}{L}
\sum_{j=0}^L \mathcal{L}_j(x_1,x_t,t;h),
\]
where (from \Cref{eq:dfm:loss})
\[
\mathcal{L}_j(x_1,x_t,t;h)
=
-\,\bar g_{t,h}\!\left[
p_{1|t}(x_t^j \mid x_t) - \delta_{x_1^j}(x_t^j)
+ \bigl(1-\delta_{x_1^j}(x_t^j)\bigr)\log p_{1|t}(x_1^j \mid x_t)
\right],
\]

\textbf{Budget-aware blending.}
With a threshold $\tau$ on the step size, blend per sample $b$ such that tiny steps $(h<\tau)$ optimize the DFM path loss $\mathcal{L}_{\text{dfm}}$, while larger steps distill to the shortcut teacher.
\begin{equation}\label{eqn:loss:blend}
m_b \;=\; \mathbb{I}[\,h_b < \tau\,],\qquad
\mathcal{L} \;=\; \frac{1}{B}\sum_{b=1}^{B}\!\Big( m_b\,\mathcal{L}^{(b)}_{\text{dfm}} \;+\; (1-m_b)\,\mathcal{L}^{(b)}_{\text{dist}} \Big).
\end{equation}

\section{Experiments}

\subsection{Experimental Setup}
\label{sec:exp:setup}

\paragraph{Training.}
\label{sec:exp:pretrain}
Step-aware DFM training incurs additional model evaluations per batch (for shortcut RK-2/RK-4 teachers), making from-scratch optimization expensive. We therefore adopt a \textbf{pretrain \(\rightarrow\) fine-tune} protocol: first pretrain a plain DFM backbone, then fine-tune it with step-aware objectives and the cumulative scalar update. We pretrain the DFM model following \citet{NEURIPS2024_f0d629a7} and cover two source distributions—\emph{uniform} and \emph{mask}—and three model sizes: 0.169B, 1.3B, and 1.7B parameters for the uniform source, plus a 0.169B mask-source model. 
These checkpoints serve as initialization for \ourmethod{} fine-tuning. \Cref{sec:app:imp:pre_model_arch} contains all architecture and training details.

\label{sec:subsec:data:token}
We train on FineWeb-Edu~\citep{lozhkov2024fineweb-edu} and evaluate on WikiText-103~\citep{merity2016pointer}. 
We use GPT-2 tokenizer and, during preprocessing, we append an EOS token to each document and pack the resulting token stream into contiguous blocks of length 1\,024. We concatenate shorter samples to reach the desired 1\,024 token target.

\label{seq:exp:paragraph:step_size_sch}
\textbf{Step–size schedule.} Shortcut integration (RK-2/RK-4) requires evaluating the model at both $h$ and $h/2$ inside training step. 
To cover a broad range of inference budgets with a single model, we sample $h$ from a logarithmic grid $h \in \lbrace 2^{k} \rbrace$ for $k\in \lbrace -10, -9, -8, \ldots, -1, 0 \rbrace$. The grid covers everything from tiny steps for precise path following to very large steps for few-step generation. In each minibatch, we sample \(h\) from the grid (uniform over values unless stated otherwise); the shortcut teacher then internally uses the required \(h/2\) sub-evaluations. We report variants of the \(h\)-sampling policy and their impact in \Cref{sec:subsec:how:step:size}. For budget-aware blending (\Cref{eqn:loss:blend}), we set the threshold \(\tau=2^{-9}\) for simplicity.

\textbf{Scheduler choice $\kappa$.}
There are many valid probability–path schedulers in DFM (e.g., convex or cosine).
For simplicity, we use the linear scheduler $\kappa(t)=t$, which yields
$g(t)=\dot\kappa(t)/(1-\kappa(t))=1/(1-t)$ and the cumulative scalar
$\bar g_{t,h}=\tfrac{1}{h}\ln\!\big(\tfrac{1-\kappa(t)}{1-\kappa(t+h)}\big)=\tfrac{1}{h}\ln\!\big(\tfrac{1-t}{1-t-h}\big)$.
This choice keeps the path well–behaved, simplifies implementation, and focuses our study on step–aware training rather than scheduler design.

\textbf{Shortcut Model.}
We make the generator \emph{step–aware} by conditioning on the intended step size $h$ alongside time $t$, i.e., $\ell=\theta(x_t,t;h)$.
During training we sample $h$, construct a shortcut teacher over $[t,t{+}h]$ using an EMA copy of the student, and apply the budget-aware losses from \Cref{sec:method:training}. 
During inference, we fix a budget of $S\in \lbrace 2^{k} \rbrace$ steps and $k\in \lbrace 0, 1, 2, \ldots, 9, 10 \rbrace$, set $h=1/S$, and run the step-aware sampler. 
Implementation details appear in \Cref{app:shortcut-model}.

\textbf{Measurement.}
We report several complementary metrics:
\emph{Perplexity (PPL)} measured by a fixed reference LM (\texttt{gpt2-large}); lower is better;
and \emph{Entropy} – the average uncertainty of our model’s token distributions; lower indicates sharper, more decisive predictions; 
\emph{Token accuracy}: the fraction of model prediction that match the ground-truth; higher is better; and MAUVE~\citep{pillutlaMAUVEMeasuringGap2021}, divergence-based metric that measures how similar the distribution of text generated by a model is to that of real human text.
All metrics are computed on the evaluation split and averaged across sequences.

% \textbf{Baselines.}
% We compare \ourmethod with two diffusion language models: LLaDA-8B~\citep{nieLargeLanguageDiffusion2025} and Dream-7B~\citep{yeDream7BDiffusion2025}, each in \emph{Base} and \emph{Instruct} variants, and against DFM baselines at three model sizes (0.169B, 1.3B, 1.7B). All baselines use the same evaluation setup.

\textbf{Baselines.}
We compare \ourmethod to a broad set of discrete diffusion and few-step generative models. In the main text, we report results against two diffusion language models, LLaDA-8B~\citep{nieLargeLanguageDiffusion2025} and Dream-7B~\citep{yeDream7BDiffusion2025}, each in \emph{Base} and \emph{Instruct} variants, as well as discrete flow-matching (DFM) baselines at three model sizes (0.169B, 1.3B, 1.7B). To further contextualize performance,~\Cref{app:subsec:expanded-baselines} expands this comparison to include multi-round refinement models such as SDTT~\citep{deschenaux2024beyond}, hybrid discrete–continuous diffusion models such as HDLM~\citep{fathi2025unifying}, and several masked-diffusion / re-masking systems (MDLM~\citep{sahoo2024simple}, SEDD~\citep{lou2023discrete}, ReMDM~\citep{wang2025remasking}), all evaluated under a fixed few-step generation budget.

\begin{table*}
	\centering
	\small
	\tabcolsep 4.5pt
		\caption{Ablation on scaler formulation across NFEs: integrating the scheduler within each step (Cumulative Scalar) yields a closed-form, probability-preserving update that sharply lowers GPT-2 perplexity—especially at 1--2 NFEs—while maintaining comparable entropy.}
	\label{tab:perp:diff:velocity}
	\scalebox{1.0}{
		\begin{tabular}{c|rr|rr|rr|rr|rr}
			%\toprule
			 & \multicolumn{2}{c|}{1} & \multicolumn{2}{c|}{2} & \multicolumn{2}{c|}{4} & \multicolumn{2}{c|}{8} & \multicolumn{2}{c}{1\,024} \\
			\multirow{-2}{*}{Solver} & ppl.                          & ent.                      & ppl.                         & ent.                       & ppl.                         & ent.                       & ppl.                         & ent.                       & ppl.                         & ent.                         \\  \hline
			
			Scaler                                         & 1\,312.65                      & 6.45                     & 462.31                      & 6.42                      & 194.29                      & 6.90                      & 97.51                       & 7.16                      &    85.61                         &               7.84              \\
			\rowcolor[HTML]{DAE8FC} 
			Cum. Scalar                                 & 514.40                       & 6.08                     & 333.07                      & 6.60                      & 176.19                      & 6.97                      & 90.49                       & 7.29                      &    87.36                        &     7.91                         \\
			
			% \bottomrule
		\end{tabular}
	} 

\end{table*}

\subsection{Results}
\label{seq:res:experimental}

\paragraph{Cumulative Scalar Improves Few-Step Sampling.}
\Cref{tab:perp:diff:velocity} shows  RK-4 + Cumulative Scalar performs better than only RK-4. Consistent gains at small budgets: GPT-2 perplexity drops by \textbf{60.8\%} at \textbf{1 NFE} (\(1\,312.65 \rightarrow 514.40\)), \textbf{28.0\%} at 2, \textbf{9.3\%} at 4, and \textbf{7.2\%} at 8. The improvement comes from integrating the scheduler’s rate over \([t,t{+}h]\) (\Cref{eq:sample:average:velocity}), which better matches a single large step to the effect of many small steps and reduces discretization bias. As \(h\) shrinks (more NFEs), the gap narrows but Cumulative Scalar remains superior. Entropy stays comparable and is slightly higher at larger budgets (e.g., \(7.29\) vs.\ \(7.16\) at 8 NFEs), indicating preserved diversity alongside lower perplexity.

\begin{figure}
    \centering
    \includegraphics[width=0.92\linewidth]{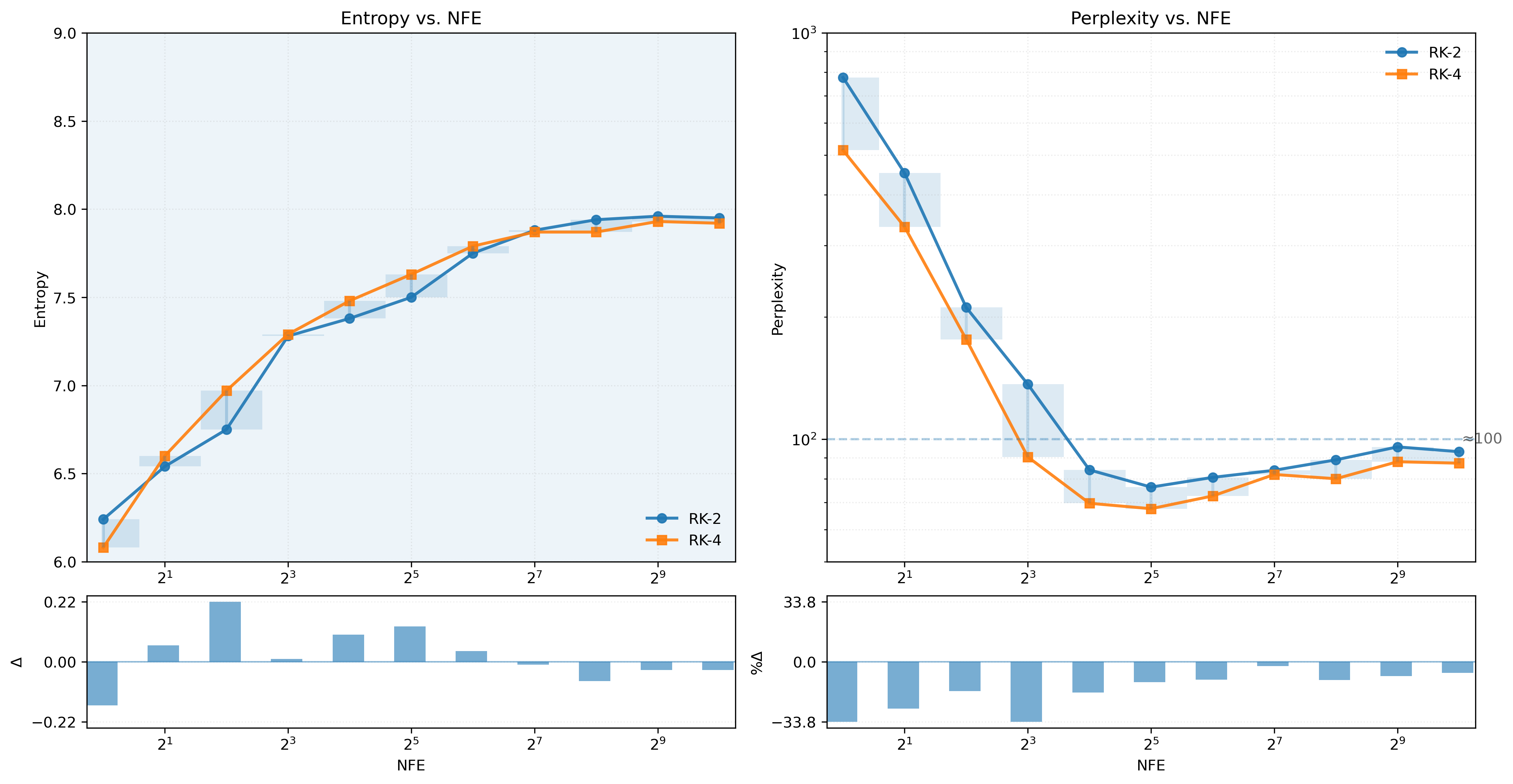}
    \vspace{-4mm}
    \caption{\small RK-2 vs. RK-4 across NFE. Top panels show entropy (linear y) and perplexity (log–log), with ribbons and vertical connectors highlighting pointwise gaps. The bottom shows the deltas ($\Delta$ entropy = RK-4 $-$ RK-2; \%$\Delta$ perplexity = RK-4/RK-2 $-1$). RK-4 has consistently lower perplexity ratio $\simeq$ $0.88 \times$ with a small entropy trade-off (median 7.63 vs. 7.5), making it the stronger choice for generation over most NFE settings.}
    \label{fig:rk_4_vs_heun}\vspace{-3mm}
\end{figure}

\textbf{The effect of RK-4 compared to RK-2 (Heun average)} is shown in 
\Cref{fig:rk_4_vs_heun}. All evaluations use Cumulative Scalar.
Overall, RK-4 performs better for generation because it delivers lower perplexity (typical/median  $\simeq$ 80 vs. RK-2's $\simeq$ 84), especially at lower NFEs. RK-4 generated a better result with a huge cap at fewer steps, by increasing the value of steps, the cap will decrease, but still, RK-4 is better. The trade-off is a slightly higher entropy (median $\simeq$ 7.63 vs. 7.50), but the perplexity improvement is larger and more relevant for text quality. Using the median to compare methods is helpful because both metrics (especially perplexity) can be skewed by occasional spikes at certain NFEs. \Cref{app:more_results} discusses RK-4 and RK-2's training-inference time trade-offs, includes an investigation into how step-size weights shape few-step fidelity, and the effect of source distribution.

\textbf{\ourmethod vs.\ DFM performance across scales} is shown in \Cref{fig:fdm_vs_fs-fdm} which compares the two methods at three parameter sizes. Across all sizes, \ourmethod reaches the strong-quality regime in far fewer function evaluations: perplexity drops sharply, accuracy saturates early, and entropy converges to the same range as many-step DFM. The gold markers at NFE=8 highlight that \ourmethod already operates near DFM's quality plateau after only a few steps, while DFM requires substantially more steps to reach the same values. To measure accuracy across steps and sizes, we randomly change $50\%$ of the tokens in each sequence with other tokens, provide the remaining $50\%$ as context, and measure the ability to predict the changed tokens. \ourmethod consistently matches or surpasses DFM with dramatically fewer evaluations. This behavior is consistent between all evaluated model sizes. As NFE increases (i.e., step size $h \to 0$), \ourmethod smoothly converges to DFM: the cumulative scalar factor satisfies $\bar g_{t,h} \to g(t)$, so the updates coincide and the curves merge. The slight accuracy dip at very large NFEs likely reflects train–test mismatch: the model and shortcut teacher are optimized for large $h$, so many small steps accumulate discretization/renormalization bias and under-use the step-aware conditioning.

\textbf{Comparison to State-of-the-Art Diffusion LMs.}
\label{subsec:sota_compare}
\Cref{tab:perp:diff:fs_dfm:sota} compares \ourmethod with contemporary base diffusion LMs in the few-step regime (\(1\!\to\!16\)). Even though the largest \ourmethod model is more than four times smaller than LLaDA and Dream, the comparison illustrates \ourmethod's generation quality. \ourmethod reaches a strong-quality regime in \emph{few steps} across model sizes: perplexity drops rapidly as steps increase while entropy remains well-behaved, indicating calibrated predictions without long iterative trajectories. In contrast, LLaDA and Dream are sensitive to step count and require many more denoising steps and do not reach comparable perplexity. Even after 16 steps, LLaDA and Dream achieve MAUVE scores equivalent to our smallest models score after a single step, despite being 41 times larger. A quick look at LLaDA and Dream's generations shows the models just repeat some tokens many times in a few-step setting (see \Cref{fig:appenxi:full_demo-1}). Full result is available in \Cref{app:subsec:full:comparison}.

\begin{table*}
	\centering
	\small
	\tabcolsep 4.5pt
	\caption{\textbf{FS-DFM vs.\ diffusion LMs across step budgets.}
		Each method receives a 512-token prefix; metrics are computed on the 512-token continuation only (ppl = perplexity, ent = entropy, MVE = MAUVE~\cite{pillutlaMAUVEMeasuringGap2021}). \ourmethod\ attains competitive quality in few steps with stable entropy across sizes, whereas baselines generally require many steps.}
	\label{tab:perp:diff:fs_dfm:sota}
	\scalebox{0.77}{
		\begin{tabular}{lr|rrr|rrr|rrr|rrr|rrr@{}}
			 & Size
			&  \multicolumn{3}{c|}{1} & \multicolumn{3}{c|}{2} & \multicolumn{3}{c|}{4} & \multicolumn{3}{c|}{8} & \multicolumn{3}{c}{16} \\
			Method & (B) & ppl. & ent. & MVE & ppl. & ent. & MVE & ppl. & ent. & MVE & ppl. & ent. & MVE & ppl. & ent. & MVE \\
			\midrule
			\rowcolor[HTML]{FFFFFF} 
			Dream   & 7.00   &      1\,163.08 &        \cellcolor[HTML]{FFCCC9}1.55  & 0.005      &        785.87 &      \cellcolor[HTML]{FFCCC9}1.43& 0.005         &     752.11       &      \cellcolor[HTML]{FFCCC9}1.53 & 0.005     &           739.40     &       \cellcolor[HTML]{FFCCC9} 1.74 & 0.006        &      630.30       &         \cellcolor[HTML]{FFCCC9}2.31 & 0.005      \\
			\rowcolor[HTML]{FFFFFF} 
			LLaDA  &  8.00 &  256.07   & \cellcolor[HTML]{FFCCC9}0.84 & 0.005 &  290.35     &   \cellcolor[HTML]{FFCCC9}0.59  & 0.005  &  495.17 & \cellcolor[HTML]{FFCCC9}0.47 & 0.005&  441.26  & \cellcolor[HTML]{FFCCC9} 0.42 & 0.005 &  432.65    &   \cellcolor[HTML]{FFCCC9}0.50 & 0.005 \\
			\rowcolor[HTML]{ECF4FF} 
			\textbf{\ourmethod}                                           & 0.17                                                    & 173.39                                  & 7.67 & 0.006          & 143.77                                  & 7.85 & 0.008         & 97.07                        & 7.89                  & 0.053   & 75.78                        & 7.95  & 0.270                   & 67.42                        & 7.97                 &  0.390    \\
			\rowcolor[HTML]{ECF4FF} 
			\textbf{\ourmethod}                                           & 1.30                                                    & 231.89                                  & 7.88 & 0.006          & 169.99                                  & 7.97    & 0.013      & 99.79                        & 7.98        & 0.120             & 70.97                        & 8.01         & 0.480            & 59.84                        & 7.99           & 0.583          \\
			\rowcolor[HTML]{ECF4FF} 
			\textbf{\ourmethod}                                           & 1.70                                                    & 191.20                                  & 7.67 & 0.007          & 155.01                                  & 7.93       & 0.014   & 101.20                       & 8.03             & 0.083        & 72.84                        & 8.07        & 0.311             & 61.67                        & 8.06         & 0.550            \\
%			\bottomrule
		\end{tabular}
	} 
\end{table*}

\section{Discussion and Conclusion}

We introduced \ourmethod, a step-aware discrete flow-matching language model that matches the perplexity of a similar-size  1\,024-step discrete-flow baseline in 1\,024 tokens generation with just 8 steps, yielding up to 128 times faster sampling. The key ideas in \ourmethod are (i) conditioning on a user-specified step budget and training such that one large move agrees with many small ones, and (ii) operating at the interval level via a cumulative scalar update that preserves the probability simplex and remains stable at large steps. We realized these ideas with shortcut teachers built inside the DFM framework and stabilized by an EMA teacher. Together, these components make few-step sampling accurate, controllable, and robust.  While our experiments used the Runge-Kutta methods to approximate solutions to the Kolmogorov equations, it would be interesting to explore other ODE solvers to determine how they might affect either generation performance or training efficiency. Finally, because research in one/few-step diffusion generation is hindered by a lack of public discrete flow-matching models, we release our DFM and FS-DFM models and code. \Cref{sec:lim:future} presents more information about future work and limitations.

% \paragraph{Reproducibility Statement:}
% We document model architectures, datasets, preprocessing, and training protocols in \cref{sec:exp:setup} and \cref{sec:app:imp:detail}; evaluation metrics, ablations, and sample outputs in \cref{seq:res:experimental,app:more_results}. Theoretical details (step-aware DFM, cumulative scalar, shortcut teacher) appear in \Cref{appendix:preliminaries,app:supp-methods}. The code and checkpoints will be published soon.

% \section*{Acknowledgement}
% We would like to acknowledge Ruixiang Zhang for providing valuable insights in our initial draft and helping to address them.

\section*{Acknowledgement}
We would like to thank Ruixiang Zhang for his valuable insights and continued support throughout the project, particularly for his thoughtful feedback and guidance in addressing key challenges.

\bibliography{iclr2026_conference}
\bibliographystyle{iclr2026_conference}

\clearpage
\appendix

\part*{Appendix}
\addcontentsline{toc}{part}{Appendix}

\etocsettocstyle{\section*{Appendix Contents}}{}
\etocsetnexttocdepth{subsubsection}
\localtableofcontents

\newpage

\section{Preliminaries}
\label[appendix]{appendix:preliminaries}
\subsection{Notation}
\label[appendix]{appendix:notation}

We set the following notations:
\begin{itemize}
    \item $L$ denotes the length of sequences,
    \item $V$ denotes the \emph{vocabulary}, i.e. the set of token ids, so $|V|$ denotes the vocabulary size,
    \item For a sequence $x$ and integer $1\leq i \leq L$, $x^i$ denotes the value of $x$ at position $i$. So $x = (x^1, \ldots, x^L)$,
    \item For sequences $x$ and $y$ we set
        \[
            \delta_y(x) = \prod_{i=1}^L\delta_{y^i}(x^i)
        \]
        where $\delta_{y^i}$ denotes the usual delta function,
    \item For a sequence $y$ and  $x^i$, we set $\delta_y(x^i) = \delta_{y^i}(x^i)$,
    \item for sequences $x$ and $y$, we set $\delta_y(\overline{x^i}) = \delta_{\overline{y^i}}(\overline{x^i}) := \prod_{j\neq i} \delta_{y^j}(x^j)$
\end{itemize}

\subsection{Basics of Continuous Time Markov Chains}

We include an overview of the theory of CTMCs focusing on the time-inhomogeneous case and where the state space $S$ is discrete, since that is the type of Markov chain modeled in this paper. 

A CTMC is a stochastic process $(X_t)_{0\leq t\leq 1}$ indexed by a continuous variable $t$ that satisfies the \emph{Markov property}: for a sequence $0\leq t_1 < t_2 < \cdots < t_n \leq 1$ and elements $i_1, \ldots, i_n \in S$,
\[
\mathbb{P}(X_{t_n} = i_n | X_{t_1} = i_1, \ldots , X_{t_{n-1}} = i_{n-1}) = \mathbb{P}(X_{t_n} = i_n | X_{t_{n-1}} = i_{n-1}).
\]
The evolution of a CTMC is dictated by its \emph{transition matrices} $P(s, t)$ for $ 0\leq s \leq t \leq 1$ where for state $x$ and state $y$ in $S$,
\[
P_{x, y}(s, t) = \mathbb{P}(X_{t} = y | X_s = x).
\]
We can derive the \emph{Chapman-Kolmogorov} equations,
\begin{equation}\label{eqn:chapman-kolmogorov}
    P(s, t) = P(s, r)P(r, t)
\end{equation}
for $s \leq r \leq t$. Note that $P(s, t)$ is a stochastic matrix.

Assuming that $P$ is differentiable in $s$ and $t$, we can define \emph{infinitesimal generator} or \emph{generator} as
\[
u_t := \lim_{h\to 0} \frac{P(t, t+h) - I}{h}
\]
where $I$ denotes the identity matrix. For states $x$ and $y$, $u_t(x, y)$ denotes the $(x, y)$-entry of $u_t$. The generator and the transition probabilities also satisfy \emph{Kolmogorov forward equation}
\begin{equation}\label{eqn:kolmogorov-forward}
    \partial_t P(s, t) = P(s, t)u_t
\end{equation}
and the \emph{Kolmogorov backward equation}
\begin{equation}\label{eqn:kolmogorov-backward}
    \partial_s P(s, t) = -u_sP(s, t)
\end{equation}

\begin{prop}\label{prop:app-q-matrix-conditions}
Let $u_t$ be the infinitesimal generator of a CTMC, then $u_t$ satisfies the following conditions.
    \begin{enumerate}
    \item $u_t(x, y) \geq 0$  for all $x \neq y$ and $t$,
    \item $\sum_y u_t(x, y) = 0$ for all $x$ and $t$,
    \item $u_t(x, x) \leq 0$.
\end{enumerate}
\end{prop}

The entries $u_t(x, y)$ are the rate of change in probability from state $x$ into $y$ when $x \neq y$, while $-u_t(x, x)$ is the rate of change in probability for staying in state $x$. In other words, we have:
\begin{equation}\label{eqn:appendix-probability-update-velocity}
    \mathbb{P}(X_{t+h} = y | X_t = x) = \delta_x(y) + h u_t(y, x) + o(h).
\end{equation}
That is, $u_t$ provides a first-order Taylor approximation of the transition probabilities. This suggests that we could also construct a Markov chain if we only had the instantaneous rate of change provided by the generator $u_t$.

Given a continuous generator $u_t$ that meets the conditions in \Cref{prop:app-q-matrix-conditions}, one can define a Markov chain by defining $P(s, t)$ to be the unique solution to the forward and backward Kolmogorov equations. Equivalently, (assuming $u_t$ is integrable), we can write
\[
P(s, t) = I + \int_s^t P(s, u)u_t du.
\]
In practice, solving for $P$ is difficult, so one usually uses numerical methods to solve these ODEs.

\subsection{Jump Chains}
\label[appendix]{app:sec:jump:chains}

Given a CTMC over a discrete state space $S$ with generator $u_t$ we described how the transition probabilities can be derived from $u_t$ for small differences (\Cref{eqn:appendix-probability-update-velocity}). We show how this is used practically when sampling for DFM. Assume $h$ is small (for example 1/1\,024). The \emph{exit rate} at time $t$ for state $x$ is given by the sum
\[
\lambda_x(t) := \sum_{x \neq y}u_t(x, y) = -u_t(x, x)
\]
Since the state space is discrete, the stochastic processes jumps from one state to another suddenly, and is constant the rest of the time. For small $h$, we can assume that the velocity does not change (i.e. we can treat $u_t$ as constant on $[t, t+h]$). Under this approximation, the holding time until the next jump is approximately exponentially distributed with rate $\lambda_x(t)$. 

In this case, the chance of jumping state can then be modeled by the Poisson distribution $1-\exp(-h\lambda_x(t))$. In other words,
\[
    \mathbb{P}(\text{jump in }[t, t+h] | X_t = x) = 1 - \exp(-h\lambda_x(t)).
\]
Note that taking the Taylor expansion for the RHS shows that 
\[
    \mathbb{P}(\text{jump in }[t, t+h] | X_t = x) = h \lambda_x(t) + o(h) = \sum_{y \neq x} \delta_x(y) + hu_t(x, y) + o(h)
\]
the last term being derived from \Cref{eqn:appendix-probability-update-velocity}, justifying that the probability of a jump in $[t, t+h)$ is approximately exponentially distributed. Finally, if a jump occurs, the probability of transitioning from state $x$ to state $y$ is given by 
\[
    \mathbb{P}(X_{t+h} = y | \text{jump occurs}, X_t = x) = \frac{u_t(x, y)}{\lambda_x(t)}
\]

Specializing to the case of DFM, we can consider the factorized velocities $u_t^i$ for each token position $i$ (\eqref{eqn:factorized-velocities}). For fixed position $i$ and state $z$. The \emph{exit rate} is similarly defined as
\begin{equation}
\lambda_t^{i}(z) := \sum_{a\neq z_i}u^{\,i}_t(a, z)
=\frac{\dot\kappa(t)}{1-\kappa(t)}\Big(1-p_{1\mid t}(z_i\mid z)\Big).
\label{eq:intensity}
\end{equation}
The jump probability in token position $i$ is also approximately exponentially distributed according to the exit rate,
\begin{equation}
\mathbb{P}[\text{jump in }[t,t{+}h))=1-\exp\!\big(-h\,\lambda_t^{i}(z)\big).
\label{eq:bernoulli-jump}
\end{equation}
Conditioned on a jump, the next token is sampled from the off-diagonals, i.e. the posterior renormalized to exclude the current token:
\begin{equation}
\mathbb{P}\!\big[X^{\,i}_{t+h}=a\mid \text{jump},X_t=z\big]
=\frac{u^{\,i}_t(a,z)}{\lambda_t^{i}(z)}
=\frac{p_{1\mid t}(a\mid z)}{1-p_{1\mid t}(z_i\mid z)}\quad(a\neq z_i).
\end{equation}

\section{Supplementary Methods}
\label[appendix]{app:supp-methods}

\subsection{Sampling}
\label[appendix]{sec:method:sampling}
At test time, we simulate the \emph{jump process} (see \cref{app:sec:jump:chains}). Choose a budget \(S\) with grid \(t_s\) and steps \(h_s=t_{s+1}-t_s\) (uniform \(h_s{=}1/S\)). Initialize \(X_{t_0}\!\sim p_0\) (uniform or mask). At step \(s\), compute logits \(\ell_s=\theta(X_{t_s},t_s;h_s)\) and set
\[
p_{1|t_s}(\cdot\mid X_{t_s}) \;=\; \mathrm{softmax}(\ell_s/T).
\]
Using the \textbf{Cumulative Scalar} \(\bar g_{t_s,h_s}\) from \Cref{eq:sample:average:velocity}, form the per-position exit rate (cf. \Cref{eq:intensity})
\[
\lambda^{\,i}_s \;=\; \bar g_{t_s,h_s}\,\Big(1 - p_{1|t_s}\!\big(X^{\,i}_{t_s}\mid X_{t_s}\big)\Big).
\]
Draw a jump with the exponential holding-time law (\Cref{eq:bernoulli-jump})
\[
J^{\,i}_s \sim \mathrm{Bernoulli}\!\left(1-e^{-h_s \lambda^{\,i}_s}\right),
\]
and, if \(J^{\,i}_s{=}1\), sample the next token from the off-diagonals of \(p_{1|t_s}(\cdot\mid X_{t_s})\) renormalized to exclude the current token:
\[
X^{\,i}_{t_{s+1}} \sim \mathrm{Cat}\!\left(\frac{p_{1|t_s}(a\mid X_{t_s})}{1 - p_{1|t_s}(X^{\,i}_{t_s}\mid X_{t_s})}\ \Big|\ a\neq X^{\,i}_{t_s}\right),
\quad\text{else set }X^{\,i}_{t_{s+1}}=X^{\,i}_{t_s}.
\]
This uses exactly one forward pass of \(\theta\) per step; conditioning on \(h_s\) lets a single checkpoint support different budgets \(S\).

\subsection{RK-2 Shortcut Teacher}
\label[appendix]{app:sec:rk-2-heun}
We include a lightweight \emph{RK-2 (Heun)} shortcut teacher as an alternative to RK-4. It approximates the interval-averaged logits over $[t,t{+}h]$ with just two model evaluations: one at $(x_t,t)$ and one at a midpoint state reached by a half-step jump $(t{+}h/2)$. This keeps compute low (two forward passes) while still providing stable, CTMC-consistent targets for a single large step of size $h$; see Algorithm~\ref{alg:heun-shortcut}.

In practice, RK-2 is useful when compute or memory are tight or for ablations that isolate the effect of teacher strength. While RK-4 typically offers stronger guidance for very large steps, RK-2 strikes a good accuracy–cost balance and integrates seamlessly with our step-aware DFM setup.

\begin{algorithm}[h]
\caption{Shortcut RK-2 Teacher}
\label{alg:heun-shortcut}
\begin{algorithmic}[1]
\Require tokens $x_t$, time $t$, step $h$; Model $\theta$; velocity $\mathrm{Vel}$; CTMC jumper $\mathrm{Jump}$; \textit{use\_ema} flag
\State $h' \gets h/2$;\quad $t_{\text{mid}} \gets t + h'$
\State $\theta' \gets \text{EMA}(\theta)$ if \textit{use\_ema} else $\theta$
\State $\ell_1 \gets \theta'(x_t, t; h')$\quad\ \ \ $u_1 \gets \mathrm{Vel}(\mathrm{softmax}(\ell_1), x_t, h', t)$\quad $\tilde{x} \gets \mathrm{Jump}(x_t, u_1, h')$
\State $\ell_2 \gets \theta'(\tilde{x}, t_{\text{mid}}; h')$
\State \textbf{RK-2 average:}\; $\bar{\ell} \gets \tfrac{1}{2}(\ell_1+\ell_2)$
\State \Return $\bar{\ell}$
\end{algorithmic}
\end{algorithm}

\subsection{Checkerboard Jump–Dynamics Demonstration}
\label[appendix]{app:checkerboard-demo}

\begin{figure}[ht]
  \centering
  \begin{subfigure}{0.99\linewidth}
    \includegraphics[width=\linewidth]{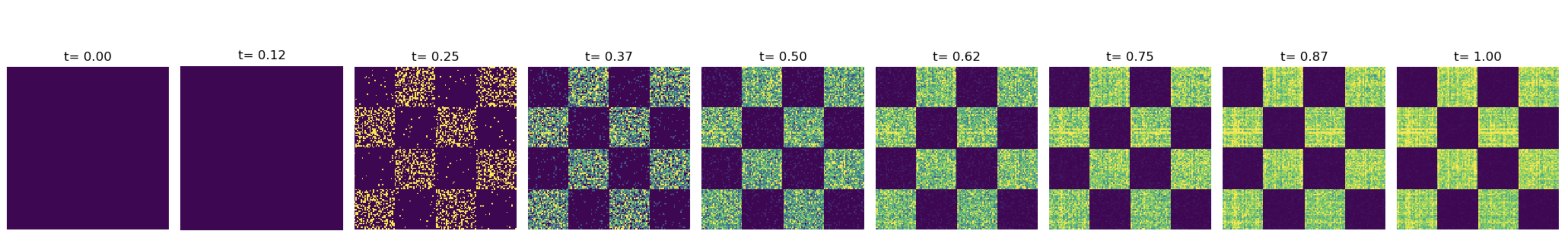}
    \caption{\small All-\texttt{[MASK]} source}
    \label{fig:checkerboard:mask}
  \end{subfigure}\hfill
  \begin{subfigure}{0.99\linewidth}
    \includegraphics[width=\linewidth]{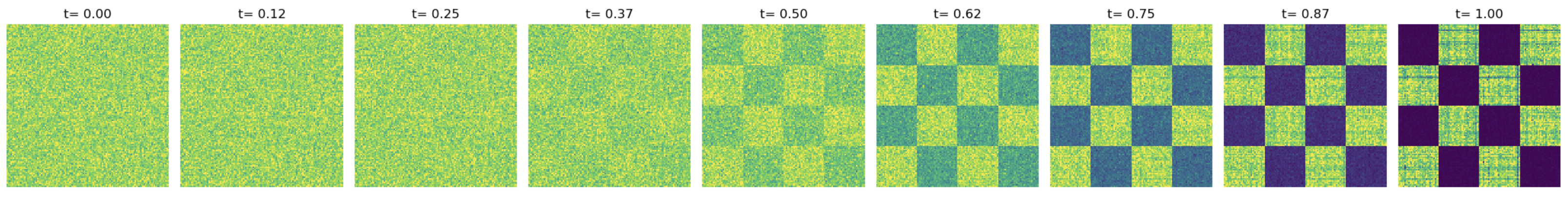}
    \caption{\small Uniform source}
    \label{fig:checkerboard:uniform}
  \end{subfigure}
  \caption{\small
  Discrete flow-matching on a \(128\times128\) checkerboard under the \emph{instantaneous} scale \(g(t)\).
  (a) With an all-\texttt{[MASK]} source, early frames exhibit almost no jumps (stalling near \(t\approx0\)).
  (b) A uniform source jumps earlier but still under-updates at tiny \(t\).
  Both patterns motivate the \emph{Cumulative Scalar} \(\bar g_{t,h}\) (\Cref{eq:sample:average:velocity}) to supply the correct probability flow over a finite step.
  }
  \label{fig:checkerboard:both}
\end{figure}

In this subsection, we want to motivate the cumulative scalar $\bar{g}_{t, h}$. We visualize discrete flow-matching on a synthetic \(128\times128\) \emph{checkerboard} target to study jump dynamics. The data generator draws the first coordinate \(x_1\) uniformly from \(\{0,\dots,127\}\); the second coordinate \(x_2\) is coupled to the parity of \(\lfloor x_1/32\rfloor\), yielding alternating \(32\times32\) blocks (checkerboard). We run a CTMC sampler on a uniform grid \(t\in[0,1]\) (100 frames) with the \emph{instantaneous} scale \(g(t)=\dot\kappa(t)/(1-\kappa(t))\), and compare two source distributions \(p_0\): (i) an all-\texttt{[MASK]} source (we add a dedicated mask token to the vocabulary), and (ii) a uniform source over tokens. We wish to emphasize that the images in \cref{fig:checkerboard:both} indicate discrete flow-matching on distributions of samples (in this case points in the plane).

The trajectories reveal the role of the scale term at early times. With an all-\texttt{[MASK]} source (\cref{fig:checkerboard:mask}), frames near \(t\approx0\) show almost no motion: common schedulers make \(g(t)\) small at the start, so exit rates are tiny and the chain stalls despite having the correct denoising direction (see the jump-process formulation in \cref{app:sec:jump:chains}). A uniform source (\cref{fig:checkerboard:uniform}) spreads initial mass over valid tokens and induces earlier movement, yet the first frames still under-shoot when driven solely by \(g(t)\). These behaviors motivate replacing the instantaneous scale by the \emph{Cumulative Scalar} \(\bar g_{t,h}\) (\cref{eq:sample:average:velocity}), which couples the update strength to both the current time \(t\) and the finite step \(h\) and makes few-/one-step updates effective.

\section{Implementation Details}
\label[appendix]{sec:app:imp:detail}

\subsection{Pre-train model architecture}
\label[appendix]{sec:app:imp:pre_model_arch}

\textbf{Backbone.}
We use a DiT-style G~\citep{Peebles2022DiT} with rotary attention. Tokens are embedded once, rotary phases are computed once per forward pass and applied to $Q,K$, and each block applies multi-head self-attention followed by an MLP. A conditioning vector $c$ drives \emph{adaptive layer normalization}: each block predicts per-channel shift/scale (and a small residual weight) from $c$ and applies them before attention and before the MLP. The final linear head is zero-initialized for stable start-up and produces logits; conversion of logits to a CTMC generator and ODE stepping are handled by the solver outside the backbone.

\textbf{Time and step-size conditioning.}
We embed continuous time $t\!\in\![0,1]$ with a sinusoidal MLP. We also embed the intended inference step size $h$ and fuse the two with a linear+SiLU layer:
\[
c=\phi_{\text{time}}(t)\quad\text{or}\quad
c=\mathrm{SiLU}\!\big(W\,[\phi_{\text{time}}(t);\phi_{\Delta t}(h)]\big).
\]
The same $c$ is fed to all transformer blocks and the final layer, making the model \emph{step-aware}: a single set of weights works with different integrator schedules at inference (few large steps or many small ones) without retraining. The backbone’s output remains logits; downstream components convert these to CTMC-valid generators and perform the PF-ODE update for the chosen step size $h$.

\subsection{Shortcut Model: Implementation Details}
\label[appendix]{app:shortcut-model}

\textbf{Training/evaluation protocol.}
In training, the student predicts $\ell=\theta(x_t,t;h)$ and is supervised by (i) a small-step path loss and (ii) a large-step distillation loss against the shortcut teacher, blended by the budget-aware rule in \cref{eqn:loss:blend}. 
At evaluation, we select a step budget $S$, set $h=1/S$, and apply the Euler-with-average-velocity update each step; no additional tuning is required.

\subsection{Training Details}
For initialization, we run 1M iterations with a global batch size of 128; each sample contributes 512 predicted tokens, for a total of 65.5B supervised tokens. For fine-tuning (same 512 predicted tokens/sample, 500K iterations per model), batch sizes are 96 (0.17B),32 (1.3B), and 16 (1.7B), yielding about 24.6B, 8.19B, and 4.10B tokens, respectively. Training used 8$\times$ A100–80GB compute nodes. We optimize all models with AdamW, using a cosine-annealed learning rate schedule. We use a peak learning rate of $3\mathrm{e}{-4}$ for pre-training and $1\mathrm{e}{-5}$ for fine-tuning.

\section{Limitations and Future Work}
\label[appendix]{sec:lim:future}

Our study primarily targets long-horizon language modeling, and our evaluation emphasizes perplexity and masked-token accuracy; broader assessments (instruction following, QA/reasoning, and multilingual settings) remain to be done. A practical limitation is the lack of public checkpoints for discrete flow-matching: Accordingly, we plan to release DFM and FS-DFM models and training code to support reproducibility and downstream research. On efficiency, we aim to learn adaptive, position-aware step schedules and budget controllers that maintain few-step quality while further reducing inference cost. Scaling to ultra-long contexts (8–32k+ tokens) with memory-efficient attention and retrieval is another important direction. Beyond likelihood metrics, we will incorporate alignment-oriented objectives (e.g., preference optimization) to improve instruction fidelity and calibration. Another choice made in our experiments was using the Runge-Kutta method RK-4 to approximate solutions to the Kolmogorov equations. It would be interesting to experiment with other ODE solvers and see how they affect performance. Also, we only inspected either uniform source or masked sources, it would be interesting to examine how a hybrid scheduler mixing masked and uniform sources could affect performance. Finally, integrating FS-DFM with systems techniques such as speculative decoding and cache reuse could compound latency gains at deployment, helping translate few-step discrete flows into practical, general-purpose generators.

\section{More Results}
\label[appendix]{app:more_results}

This section includes more experiments, especially ablation studies, and some generated samples.

\subsection{Ablation Studies}
We assess each design choice by varying a single factor while holding the rest of the pipeline fixed. Unless stated otherwise, all ablations use the same data split, model size, optimizer and schedule (AdamW with cosine), the step-size grid \(\{1/1024,\ldots,1\}\) with same policy of sampling, the shortcut teacher (RK-4 with EMA) during fine-tuning, the \emph{Cumulative Scalar} for training and sampling, a uniform source distribution, and identical inference budgets \(S\in\{1,2,4,8\}\) with temperature \(T{=}1\).

\textbf{Effect of Step-Size Sampling Weights on Few-Step Fidelity.}
\label[appendix]{sec:subsec:how:step:size}
We ablate the \emph{training-time sampling weights} over the step-size grid from \Cref{seq:exp:paragraph:step_size_sch} (ordered from the smallest to the largest step) to understand how biasing minibatch selection toward large steps affects few-step performance. We evaluate five policies: (i) a ``tail-boosted'' \textbf{(TB-10)} scheme with uniform weights except on the largest step, \([\mathbf{10}, 1,1,1,1,1,1,1,1,1,1]\); (ii) a stronger tail boost \textbf{(TB-20)}, \([\mathbf{20}, 1,1,1,1,1,1,1,1,1,1]\); (iii) pure uniform \textbf{(PU)} with the same weight on all step sizes; (iv) a geometric~\textbf{(G)}, large-step–biased schedule \([2^{10}, 2^9, \dots, 2^1, 2^0]\); and (v) an \emph{annealed geometric} \textbf{(AG)} variant that starts as geometric~\textbf{(G)} and transitions to pure uniform \textbf{(PU)} by doubling weights every $10\,000$ steps up to $2^{10}$, so by $100\,000$ steps, all weights equal $2^{10}$. This experiment probes two intuitions: (a) modest to strong emphasis on the largest \(h\) should sharpen single/few-step accuracy by exposing the student to more shortcut-distilled targets; and (b) annealing the bias away (policy AG) preserves this benefit early while restoring balanced coverage later, stabilizing path-following at small \(h\) without sacrificing few-step quality. The results (\Cref{tab:perp:diff:step_size_sampling_weights}), indicate that the strongest tail boost, \textbf{(TB-20)}, performs best. We adopt this setting for all subsequent experiments.

\begin{table*}
	\centering
	\small
	\tabcolsep 4.5pt
	\caption{Step-size weight ablation over $h \in \lbrace 2^{-k} \rbrace$ for $k\in \lbrace -10, -9, -8, \dots, -1, 0 \rbrace$: TB-10, TB-20, PU~(uniform), G~(geometric), AG~(annealed). Bias toward large \(h\) improves few-step quality.}
	\label{tab:perp:diff:step_size_sampling_weights}
	\scalebox{1}{
		\begin{tabular}{c|rr|rr|rr|rr|rr}
			%\toprule
			  & \multicolumn{2}{c|}{1} & \multicolumn{2}{c|}{2} & \multicolumn{2}{c|}{4} & \multicolumn{2}{c|}{8} & \multicolumn{2}{c}{1\,024} \\
			Policy & ppl.                          & ent.                      & ppl.                         & ent.                       & ppl.                         & ent.                       & ppl.                         & ent.                       & ppl.                         & ent.                         \\  \hline
			
			\textbf{TB-10}              &            834.85         &           6.51         &          325.32            &          6.66        &         159.59              &      6.99    &           94.72           &         7.19                                &              85.13              &            7.94               \\
			
			\rowcolor[HTML]{DAE8FC} 
			\textbf{TB-20}                        &            514.40         &       6.08             &           333.07           &        6.60          &         176.19               &       6.97     &            90.49          &       7.29                                  &              87.36              &           7.92                \\
			
			\textbf{PU}             &         3.01            &          0.64          &              30.81        &         1.73         &          18.67            &  2.31        &           21.54           &             2.73     &            262.53                &    6.96                       \\
			
			\textbf{G}         &           1\,353.99          &           7.57         &               635.59          &          7.71              &    250.89      &          7.85            &       116.99           &7.85     &     96.20   &      7.95             \\
			\rowcolor[HTML]{FFCCC9}

			\textbf{AG}     &         906.15            &         5.82           &     439.41                 &        5.93          &           161.82              &    6.37      &         87.73             &         6.63      &             77.83               &              7.76             \\
%			\bottomrule
		\end{tabular}
	}
\end{table*}

\textbf{Effect of source Distribution for Few-Step Generation.}
We study the effect of the \emph{source distribution} used by DFM on few-step sampling. Following \citet{NEURIPS2024_f0d629a7}, we compare a \textbf{mask source} (all positions start as a dedicated \texttt{[MASK]} state) against a \textbf{uniform source} (initialized with random tokens). With only one or a few steps, the CTMC has limited opportunity to escape \texttt{[MASK]}; transitions concentrate on a small set of outcomes or remain at \texttt{[MASK]}, producing collapse and high perplexity. In contrast, the uniform source spreads initial mass over real tokens, so the model performs corrective refinements rather than first-escape jumps. This yields markedly lower perplexity with healthy entropy at tiny budgets and remains competitive as the budget grows. \Cref{tab:perp:diff:distiribution} shows that uniform source is consistently superior in the few-step regime, while the performance gap narrows with larger NFE as more transitions become available. For this experiment, we used the RK-2 method for both mask and uniform sources.

\begin{table*}
	\centering
	\small
	\tabcolsep 4.5pt
	\caption{Effect of source distribution on \ourmethod. \emph{Uniform} source initialization avoids first-escape bottlenecks from \texttt{[MASK]} and delivers lower perplexity with comparable entropy at small NFEs; a \emph{mask} source often collapses in the few-step regime.}
	\label{tab:perp:diff:distiribution}
	\scalebox{1}{
		\begin{tabular}{c|rr|rr|rr|rr|rr}
			%\toprule
			Source& \multicolumn{2}{c|}{1} & \multicolumn{2}{c|}{2} & \multicolumn{2}{c|}{4} & \multicolumn{2}{c|}{8} & \multicolumn{2}{c}{1\,024} \\
			Distribution & ppl.                          & ent.                      & ppl.                         & ent.                       & ppl.                         & ent.                       & ppl.                         & ent.                       & ppl.                         & ent.                         \\  \hline
			
			Mask                                       & 4\,438.49                    & 8.62                   & 2\,844.42                      & 8.64                 & 1\,082.07                       & 8.48         & 413.43                     & 8.40                                        &    364.61                         &               8.36              \\
			\rowcolor[HTML]{DAE8FC} 
			Uniform                                 & 777.02                       & 6.24                     & 451.79                      & 6.54                      & 211.04                      & 6.75                      & 136.60                       & 7.38                      &    93.24                       &     7.95                         \\
			
%						\bottomrule
		\end{tabular}
	} 
\end{table*}

\subsection{Training–Inference Time Trade-offs}
\label{app:subsec:training-inference-time}
Shortcut teachers (RK-2/RK-4) improve few-step fidelity by querying the model at multiple abscissae within each training step, which increases per-iteration compute relative to pure DFM. In our setup, RK-2 adds roughly \textbf{33\%} wall-clock time per training step, while RK-4 adds about \textbf{100\%} (due to the extra forward passes and intermediate updates). We mitigate this cost in two ways: (i) we \emph{fine-tune} from a strong pre-trained checkpoint, limiting the number of optimization steps required; and (ii) we use shortcut teachers \emph{only during fine-tuning} to teach large-step consistency, whereas \emph{inference} uses \textbf{very few steps} (e.g., 1–8 NFEs), yielding substantial latency and energy savings over the model’s lifetime. Consistent with the ablation in \Cref{sec:subsec:how:step:size}, the stronger tail-boost policy (\textbf{\textsc{TB-20}}) delivers the best quality–compute trade-off: it concentrates training on large steps—where shortcut guidance is most valuable—without degrading small-step path following. Overall, the transient training overhead is amortized by the persistent inference gains, making the net runtime favorable for deployment.

% \textcolor{red}{\subsection{More Baselines:}}
% \label[appendix]{app:subsec:more:baselines}

\subsection{Expanded Baseline Comparison and Positioning of \ourmethod}
\label[appendix]{app:subsec:expanded-baselines}

To more comprehensively situate the performance of \ourmethod within the broader landscape of discrete diffusion and few-step generative models, we extend our baseline comparisons in \Cref{tab:appendix:diff:fs_dfm:sota:all} to include a wider set of modern architectures. This expanded set goes beyond the baselines reported in the main paper and incorporates models that vary significantly in training budgets, refinement strategies, masking mechanisms, and model size. Including these baselines is essential for establishing a more complete understanding of how \ourmethod performs relative to systems that rely on iterative refinement, large-token pre-training, or hybrid continuous–discrete generative mechanisms.

\begin{table}[t]
\centering
\small
\tabcolsep 5pt
\caption{
\textbf{Expanded baseline comparison under an 8-step generation budget.}
All models are evaluated with the same number of sampling steps (8).  
For \ourmethod{}, the first 65.5B tokens correspond to the DFM pre-training corpus, and the additional tokens (\ddag) denote fine-tuning on the FS–DFM objective.  
This table aggregates the most widely used discrete diffusion, masked-diffusion, and hybrid few-step baselines, allowing for a unified comparison across model size, training-token scale, and evaluation metrics. Training data metrics are provided to facilitate better comparison across methods.  
\label{tab:appendix:diff:fs_dfm:sota:all}}
\scalebox{0.95}{

\begin{tabular}{lrrrr}
% \toprule
\textbf{Model} & \textbf{Size (B)} & \textbf{Tokens (B)} & \textbf{PPL} & \textbf{Entropy} \\
\midrule
Training Data & - & - & 14.7 & 7.70
\\
\midrule
\multicolumn{5}{c}{\textit{Multi-round refinement models}} \\
SDTT (1 round)~\citep{deschenaux2024beyond} & 0.86 & -- & 612.38 & 8.38 \\
SDTT (3 rounds)~\citep{deschenaux2024beyond} & 0.86 & -- & 376.19 & 8.19 \\
SDTT (5 rounds)~\citep{deschenaux2024beyond} & 0.86 & -- & 220.81 & 8.09 \\
SDTT (7 rounds)~\citep{deschenaux2024beyond} & 0.86 & -- & 131.71 & 7.78 \\
% \midrule
\multicolumn{5}{c}{\textit{Hybrid discrete–continuous diffusion models}} \\
HDLM ($\epsilon=0.01$)~\citep{fathi2025unifying} & 0.16 & 524 & 279.60 & 7.54 \\
HDLM ($\gamma=0.05$)~\citep{fathi2025unifying} & 0.16 & 524 & 192.74 & 8.13 \\
% \midrule
\multicolumn{5}{c}{\textit{Masked diffusion / discrete iterative models}} \\
MDLM–OWT~\citep{sahoo2024simple} & 0.17 & 262 & 837.58 & 8.47 \\
% DUO~\citep{sahoo2025diffusion} & 0.17 & 2621.4 & 68.49 & 5.28 \\
SEDD~\citep{lou2023discrete} & 0.32 & -- & 602.09 & 8.42 \\
ReMDM–cap~\citep{wang2025remasking} & 0.17 & 262 & 559.74 & 8.34 \\
ReMDM–rescale~\citep{wang2025remasking} & 0.17 & 262 & 560.06 & 8.35 \\
ReMDM–conf~\citep{wang2025remasking} & 0.17 & 262 & 549.86 & 8.35 \\
ReMDM–loop~\citep{wang2025remasking} & 0.17 & 262 & 1358.84 & 8.41 \\
\midrule
\multicolumn{5}{c}{\textit{Original Discrete Flow Matching (DFM) baselines}} \\
DFM~\citep{NEURIPS2024_f0d629a7} & 0.17 & 65.5 & 335.64 & 8.06 \\
DFM~\citep{NEURIPS2024_f0d629a7} & 1.30 & 65.5 & 331.36 & 8.11 \\
DFM~\citep{NEURIPS2024_f0d629a7} & 1.70 & 65.5 & 444.71 & 8.18 \\
% \midrule
\multicolumn{5}{c}{\textit{Fast Version of DFM}} \\
\rowcolor[HTML]{DAE8FC} \textbf{\ourmethod} & 0.17 & $65.5 + 24.6^\ddag$ & \textbf{90.49} & 7.29 \\
\rowcolor[HTML]{DAE8FC} \textbf{\ourmethod} & 1.30 & $65.5 + 8.2^\ddag$ & \textbf{77.21} & 7.40 \\
\rowcolor[HTML]{DAE8FC} \textbf{\ourmethod} & 1.70 & $65.5 + 4.1^\ddag$ & \textbf{73.65} & 7.61 \\
% \bottomrule
\end{tabular}
}
\end{table}

A first category consists of multi-round refinement models such as SDTT~\citep{deschenaux2024beyond}, which improve quality through multiple decoding rounds. Although additional rounds consistently reduce perplexity, even seven rounds of refinement remain substantially behind the performance of \ourmethod at only eight sampling steps (see \Cref{tab:appendix:diff:fs_dfm:sota:all}). Beyond this quantitative gap, the two approaches also differ conceptually: SDTT accelerates generation by sampling from a many-step teacher and distilling its predictions, whereas \ourmethod directly learns \emph{step-size--aware finite-step Kolmogorov dynamics} via RK-based ODE integration and a cumulative scalar inside the DFM framework. In other words, \ourmethod models how probability mass should evolve under different step budgets, rather than only compressing a fixed long-run sampler, which helps explain its stronger few-step behavior.

A second category includes hybrid diffusion–autoregressive models such as HDLM~\citep{fathi2025unifying}. These models leverage continuous-time diffusion in combination with powerful sequence encoders and benefit from extremely large training sets—often exceeding 500B tokens. Despite this enormous training budget, their perplexity remains noticeably higher than that achieved by \ourmethod using a considerably smaller pre-training and fine-tuning corpus. This reinforces the training-token efficiency of our fast flow–based method.

We also include several discrete masked-diffusion and re-masking systems—ReMDM variants, and SEDD~\citep{lou2023discrete}. These approaches attempt to improve sample quality by strategically masking tokens and applying multi-pass denoising schedules. Some of these baselines rely on exceptionally large training corpora, reaching into the hundreds of billions of tokens. Yet even under such favorable conditions, their perplexity and entropy remain less competitive. By contrast, \ourmethod achieves substantially stronger results with a dramatically smaller token budget, highlighting the impact of our distillation strategy and few-step training design.

Finally, we include the original DFM models from~\citep{NEURIPS2024_f0d629a7}. These models are trained on the same 65.5B-token corpus as the backbone of our system, making them a direct and informative reference point. The results show that \ourmethod improves perplexity by more than a factor of four to five compared to these predecessors while preserving the same generation budget of eight sampling steps. This comparison isolates the contribution of our flow-distilled fast sampler from confounding factors such as model size or data scale.

Taken together, this expanded baseline analysis demonstrates that \ourmethod consistently matches or surpasses models that rely on deeper refinement loops, larger-scale diffusion procedures, masking strategies, or extensive training-token resources. The results highlight both the sampling efficiency and the training efficiency of our approach, positioning \ourmethod as one of the strongest few-step discrete generative models currently available.

\subsection{Full comparison}
\label[appendix]{app:subsec:full:comparison}

\begin{table*}
	\centering
	\small
	\tabcolsep 4.5pt
	\caption{\textbf{FS-DFM vs.\ diffusion LMs across step budgets.}
		Each method receives a 512-token prefix; metrics are computed on the 512-token continuation only. \ourmethod\ attains competitive quality in few steps with stable entropy across sizes, whereas baselines generally require many steps.}
	\label{tab:appendix:diff:fs_dfm:sota}
	\scalebox{0.78}{
		\begin{tabular}{lr|rrr|rrr|rrr|rrr|rrr@{}}
			& Size
			&  \multicolumn{3}{c|}{1} & \multicolumn{3}{c|}{2} & \multicolumn{3}{c|}{4} & \multicolumn{3}{c|}{8} & \multicolumn{3}{c}{16} \\
			Method & (B) & ppl. & ent. & MVE & ppl. & ent. & MVE & ppl. & ent. & MVE & ppl. & ent. & MVE & ppl. & ent. & MVE \\
			\midrule
			\rowcolor[HTML]{FFFFFF} 
			Dream-B  & 7.00   &      1\,163 &        \cellcolor[HTML]{FFCCC9}1.55  & 0.005      &        785 &      \cellcolor[HTML]{FFCCC9}1.43& 0.005         &     752       &      \cellcolor[HTML]{FFCCC9}1.53 & 0.005     &           739     &       \cellcolor[HTML]{FFCCC9} 1.74 & 0.006        &      630       &         \cellcolor[HTML]{FFCCC9}2.31 & 0.005      \\
			
			\rowcolor[HTML]{FFFFFF} 
			Dream-I       & 7.00            &             46\,371        &      \cellcolor[HTML]{FFCCC9}0.58  & 0.005       &               64\,281             &   \cellcolor[HTML]{FFCCC9}0.35  & 0.006          &             66\,348            &          \cellcolor[HTML]{FFCCC9} 0.20    & 0.007        &         62\,740                     &        \cellcolor[HTML]{FFCCC9}0.13    & 0.006               &             57\,694                 &          \cellcolor[HTML]{FFCCC9}0.10     & 0.006            \\ 
			
			\rowcolor[HTML]{FFFFFF} 
			LLaDA-B &  8.00 &  256   & \cellcolor[HTML]{FFCCC9}0.84 & 0.005 &  290     &   \cellcolor[HTML]{FFCCC9}0.59  & 0.005  &  495 & \cellcolor[HTML]{FFCCC9}0.47 & 0.005&  441  & \cellcolor[HTML]{FFCCC9} 0.42 & 0.005 &  432    &   \cellcolor[HTML]{FFCCC9}0.50 & 0.005 \\
			\rowcolor[HTML]{FFFFFF} 
LLaDA-I &  8.00 &  8\,677   & \cellcolor[HTML]{FFCCC9}0.53 & 0.012 &  8\,869     &   \cellcolor[HTML]{FFCCC9}0.41  & 0.014  &  9\,049 & \cellcolor[HTML]{FFCCC9}0.40 & 0.029&  9\,259  & \cellcolor[HTML]{FFCCC9} 0.40 & 0.027 &  9\,289    &   \cellcolor[HTML]{FFCCC9}0.43 & 0.048 \\

			\rowcolor[HTML]{ECF4FF} 
			\textbf{\ourmethod}                                           & 0.17                                                    & 173                                  & 7.67 & 0.006          & 143                                  & 7.85 & 0.008         & 97                        & 7.89                  & 0.053   & 75                        & 7.95  & 0.270                   & 67                        & 7.97                 &  0.390    \\
			\rowcolor[HTML]{ECF4FF} 
			\textbf{\ourmethod}                                           & 1.30                                                    & 231                                  & 7.88 & 0.006          & 169                                  & 7.97    & 0.013      & 99                        & 7.98        & 0.120             & 70                        & 8.01         & 0.480            & 59                        & 7.99           & 0.583          \\
			\rowcolor[HTML]{ECF4FF} 
			\textbf{\ourmethod}                                           & 1.70                                                    & 191                                  & 7.67 & 0.007          & 155                                  & 7.93       & 0.014   & 101                       & 8.03             & 0.083        & 72                        & 8.07        & 0.311             & 61                        & 8.06         & 0.550            \\
			%			\bottomrule
		\end{tabular}
	} 
\end{table*}

\Cref{tab:appendix:diff:fs_dfm:sota} compares \ourmethod with contemporary diffusion LMs in the few-step regime (\(1\!\to\!16\)) across both baseline \emph{and} instruct models. Even though the largest \ourmethod model is multiple times smaller than LLaDA and Dream, the comparison illustrates \ourmethod's generation quality. For each example, we supply a 512-token prefix and evaluate only the 512-token continuation. \ourmethod reaches a strong-quality regime in \emph{few steps} across model sizes: perplexity drops rapidly as steps increase while entropy remains well-behaved, indicating calibrated predictions without long iterative trajectories. In contrast, LLaDA and Dream are sensitive to step count and require many more denoising steps for comparable generation quality. For the Base models we just gave the 512 tokens without any instruction but we prefixed inputs to the instruction models with the prompt: \texttt{Complete the following text as a coherent passage. Do not repeat the given prefix, just continue it.} $+$ \texttt{given tokens}

\subsection{Sample Outputs}
\label[appendix]{app:sample_outputs}

\begin{figure}[ht]
    \centering
    \includegraphics[trim=0.5cm 0.9cm 0.5cm 0.7cm, clip,width=1.\linewidth]{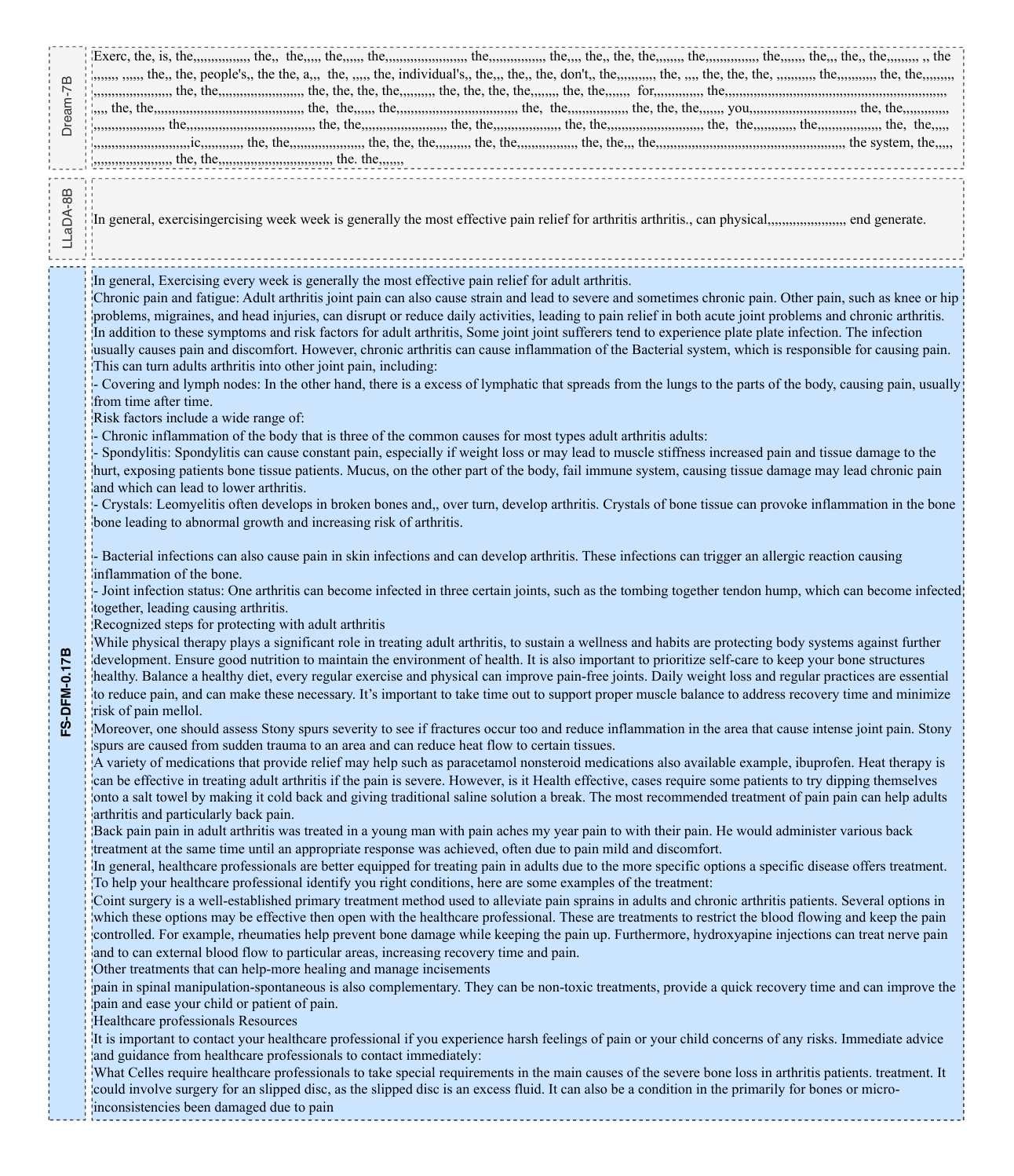}
    \vspace{-7mm}
    \caption{\small 1\,024-token unconditional generation in \emph{8 sampling steps}.
    \ourmethod (0.17B) successfully produces 1\,024 tokens under the 8-step constraint. Despite having 40x more parameters, LLaDA-8B-Instruct and Dream-7B-Instruct's 8-steps generations exhibit trailing blanks and punctuation artifacts (e.g., repeated commas).
    Generations are truncated. Complete version of \Cref{fig:demo}.}
    \label{fig:appenxi:full_demo}\vspace{-3mm}
\end{figure}

\paragraph{Overview.}
\Cref{fig:appenxi:full_demo} presents a complete, high-resolution version of the qualitative result shown in the main text (\Cref{fig:demo}), displaying the full 8-step evolution of the sample.

We would like to acknowledge Ruixiang Zhang for providing valuable insights in our initial draft, and helping to address them.

\begin{figure}[t]
    \centering
    \includegraphics[trim=0.5cm 1cm 0.5cm 0.9cm, clip,width=1\linewidth]{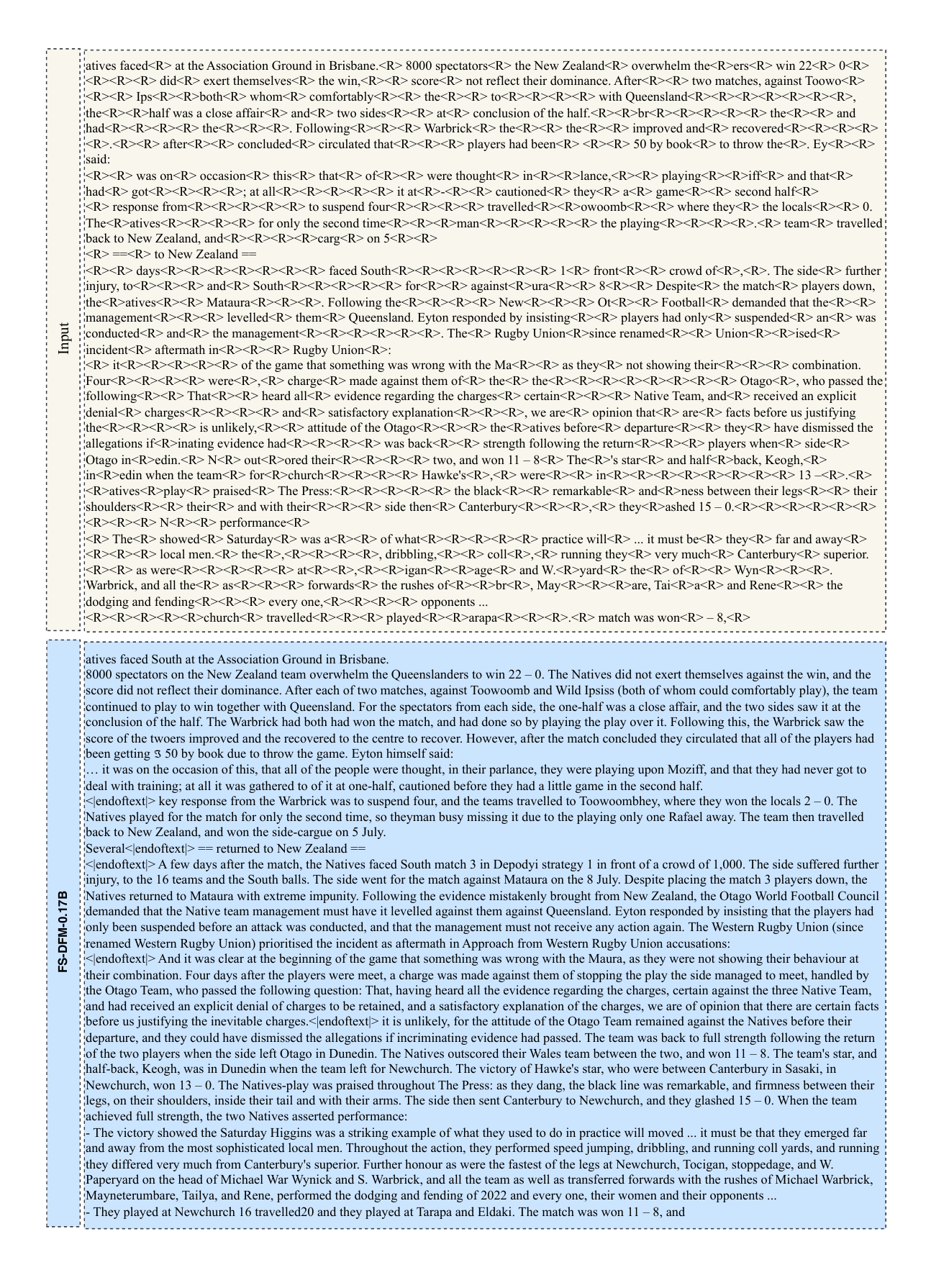}
    \vspace{-7mm}
    \caption{\small
        Corruption–recovery on a 1\,024-token passage with \textbf{50\%} random replacements (\texttt{<R>}): \ourmethod restores coherence in \textbf{8} jump steps using the Cumulative Scalar–driven CTMC sampler.
        }
    \label{fig:appenxi:full_demo-2}\vspace{-3mm}
\end{figure}

\begin{figure}[t]
    \centering
    \includegraphics[trim=0.5cm 1cm 0.5cm 0.9cm, clip,width=1\linewidth]{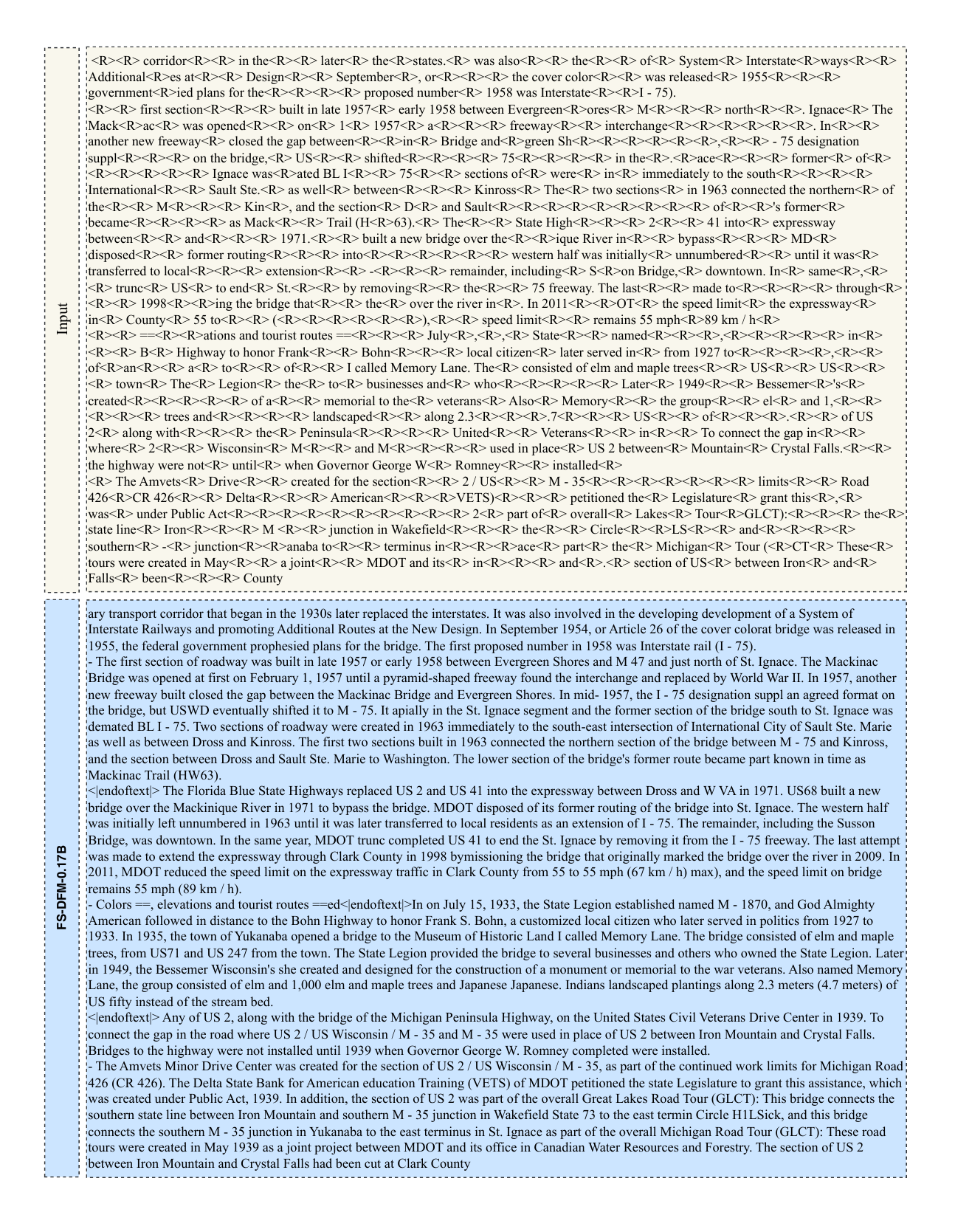}
    \vspace{-7mm}
    \caption{\small
        Corruption–recovery on a 1\,024-token passage with \textbf{50\%} random replacements (\texttt{<R>}): \ourmethod restores coherence in \textbf{8} jump steps using the Cumulative Scalar–driven CTMC sampler.
        }
    \label{fig:appenxi:full_demo-3}\vspace{-3mm}
\end{figure}

\paragraph{Corruption–recovery.}
\Cref{fig:appenxi:full_demo-2,fig:appenxi:full_demo-3} evaluate a 1\,024-token restoration task: starting from a valid sequence, we uniformly replace half of the tokens at random and ask the model to repair the corruption. \ourmethod performs this reconstruction in \textbf{8 steps} (no teacher or micro-steps at inference), demonstrating that few steps suffice to move from heavily perturbed inputs to coherent long-form text.

\begin{figure}[t]
    \centering
    \includegraphics[trim=0.5cm 1cm 0.5cm 0.9cm, clip,width=1\linewidth]{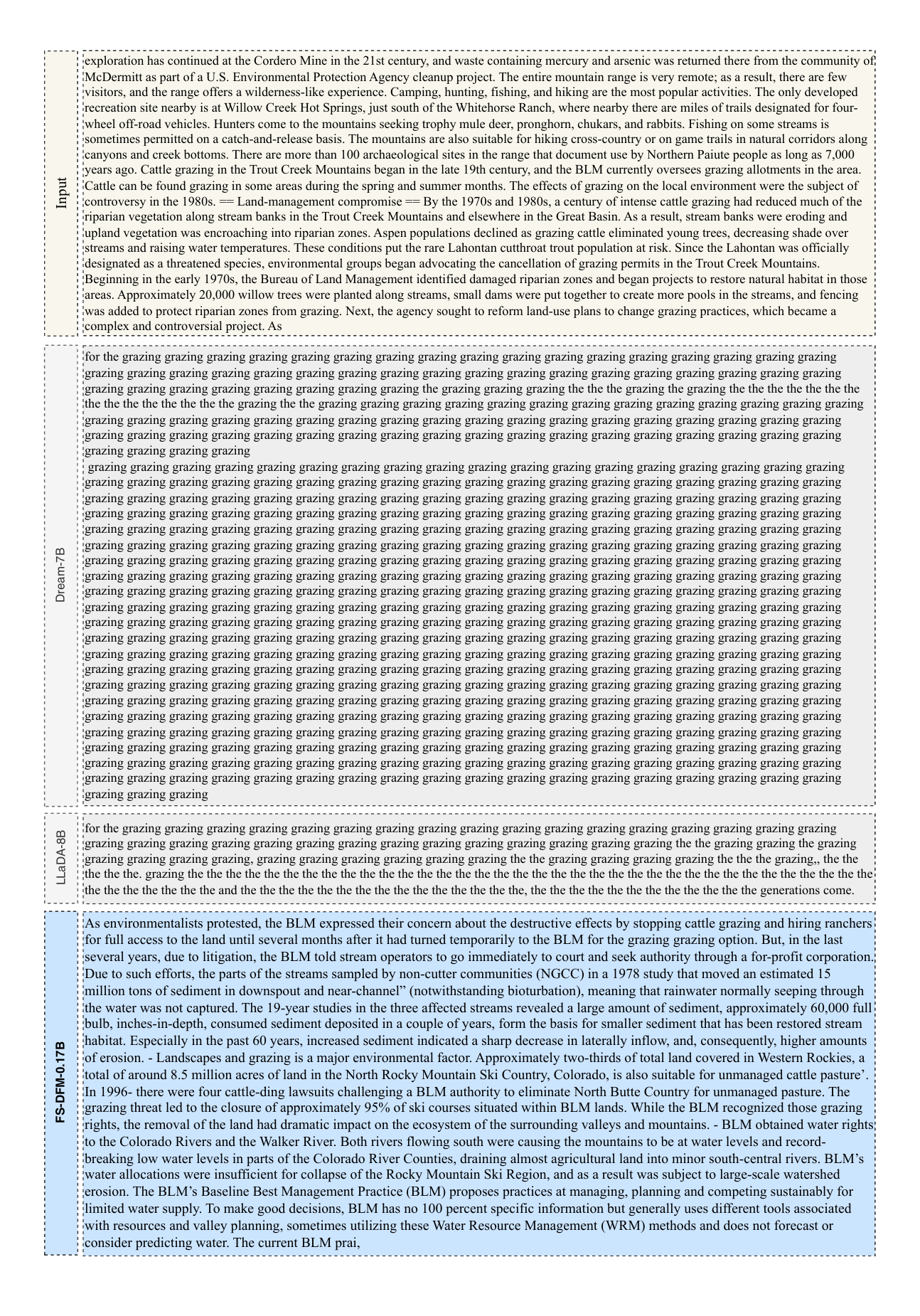}
    \vspace{-7mm}
    \caption{\small Compares \ourmethod with LLaDA and Dream on next-token continuation from a shared 512-token prefix. \ourmethod at 170 M parameters can generate coherent English on the provided topic in 8 steps while LLaDA and Dream fail to generate coherent English despite having 40x more parameters.}
    \label{fig:appenxi:full_demo-1}\vspace{-3mm}
\end{figure}

\paragraph{Head-to-head continuations (512-token prefix).}
\Cref{fig:appenxi:full_demo-1} compares \ourmethod with LLaDA and Dream on next-token continuation from a shared 512-token prefix (top panel). The subsequent panels show each method’s 8-step outputs. For LLaDA, we use \texttt{low\_confidence} remasking with block length \(\max(512/\texttt{step},\,32)\) and temperature \(0.0\).
This standardized setup highlights the qualitative differences between few-step discrete sampling (\ourmethod) and iterative remasking or diffusion-style baselines.

\begin{figure}[t]
    \centering
    \includegraphics[width=1\linewidth]{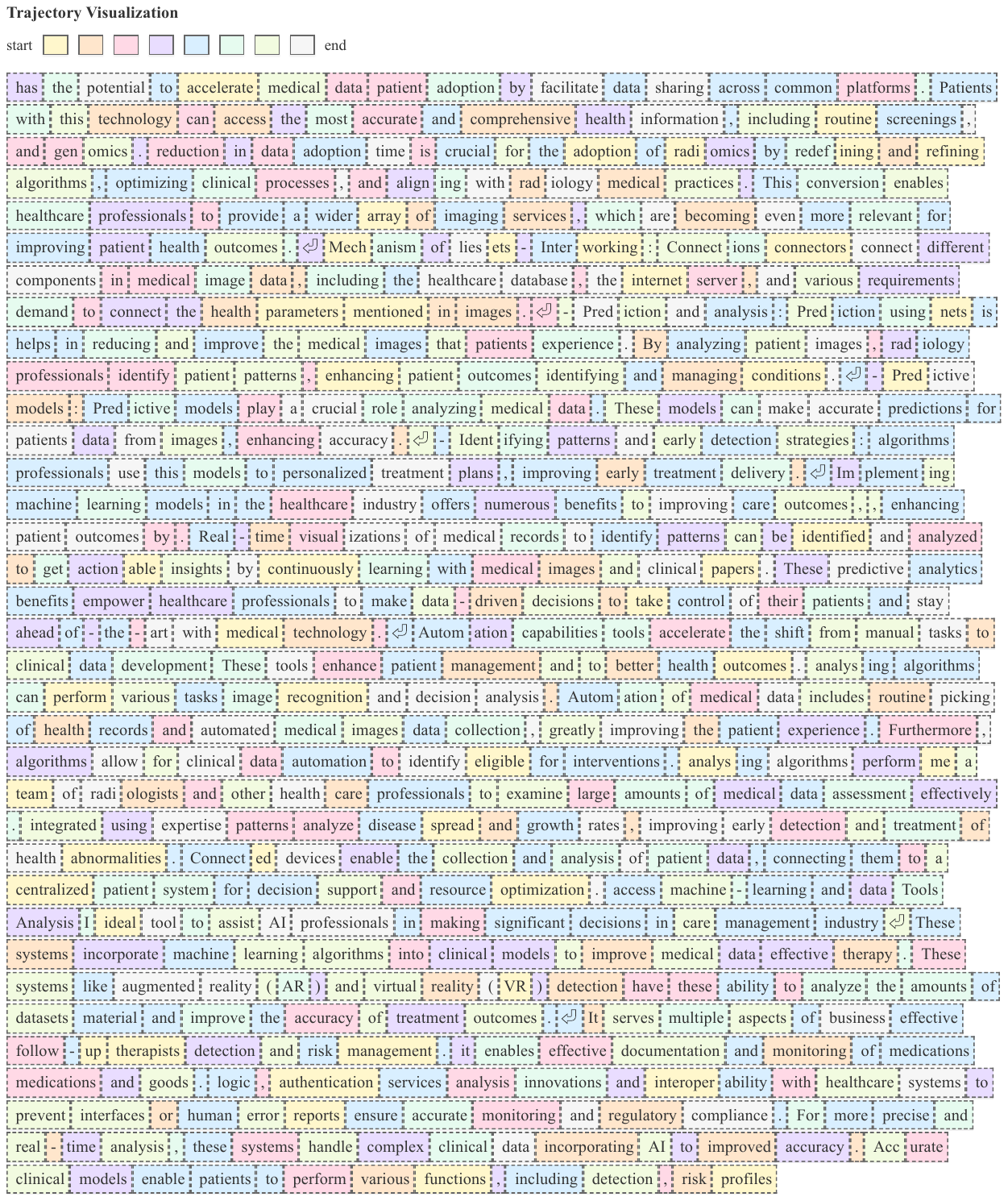}
    \vspace{-7mm}
    \caption{\small \textbf{Token-level generation timeline.} The displayed text is the final sample; the background of each token encodes the step of its \emph{last change} using eight light colors (start~$\rightarrow$~end). Early-stabilized tokens appear in early hues, while late edits trend toward end hues, making localized refinements and overall convergence easy to see. Note that many tokens are colored yellow, indicating they were predicted early in the process. This is due to the cumulative scalar (contrast with \Cref{fig:checkerboard:both}).}
    \label{fig:appenxi:full_demo_step}\vspace{-3mm}
\end{figure}

\paragraph{Trajectory visualization.}
To reveal where edits concentrate over time, \Cref{fig:appenxi:full_demo_step} colors each token by the last step at which it changed (8 bins from start\(\rightarrow\)end). Tokens that stabilize early retain an ``early'' hue, while late edits appear in ``late'' hues, making convergence and late-stage polishing patterns immediately visible.

\end{document}